\documentclass[10pt,journal,compsoc]{IEEEtran}
\usepackage{amssymb}
\usepackage{amsmath}
\usepackage{graphicx}
\usepackage{subfigure}
\usepackage{enumitem}
\usepackage{multirow}
\usepackage{color}
\usepackage{booktabs}

\usepackage[ruled,vlined]{algorithm2e}
\definecolor{commentcolor}{RGB}{110,154,155}   
\newcommand{\PyComment}[1]{\ttfamily\textcolor{commentcolor}{\# #1}}  
\newcommand{\PyCode}[1]{\ttfamily\textcolor{black}{#1}} 

\usepackage{etoolbox}

\makeatletter
\patchcmd{\@algocf@start}
  {-1.5em}
  {0pt}
  {}{}
\makeatother

\ifCLASSOPTIONcompsoc
  \usepackage[nocompress]{cite}
\else
  \usepackage{cite}
\fi
\ifCLASSINFOpdf
\else
\fi
\begin{document}
%
\title{Semantics-Guided Contrastive Network for Zero-Shot Object Detection}
%
%
%
%

\author{Caixia Yan,~\IEEEmembership{}
        Xiaojun Chang,~\IEEEmembership{}
        Minnan Luo,~\IEEEmembership{}
        Huan Liu,~
        Xiaoqin Zhang,~
        and~Qinghua Zheng~\IEEEmembership{}
\IEEEcompsocitemizethanks{
\IEEEcompsocthanksitem Caixia Yan, Minnan Luo, Huan Liu and Qinghua Zheng are with the Ministry of Education Key Lab of Intelligent Networks and Network Security, National Engineering Lab for Big Data Analytics, School of Electronic and Information Engineering, Xi'an Jiaotong University, Shaanxi 710049, China, {e-mail: yancaixia@stu.xjtu.edu.cn, \{minnluo,huanliu,qhzheng\}@mail.xjtu.edu.cn}
\IEEEcompsocthanksitem Xiaojun Chang is with the ReLER lab, AAAI, Faculty of Engineering and Information Technology, University of Technology Sydney, Australia, e-mail: xiaojun.chang@uts.edu.au.
\IEEEcompsocthanksitem Xiaoqin Zhang is with the College of Computer Science and Artificial Intelligence, Wenzhou University, China, e-mail: xqzhang@wzu.edu.cn. }
\thanks{Corresponding authors: Xiaoqin Zhang and Xiaojun Chang.}
\thanks{Manuscript received April 29 2021; revised September 12 2021, accepted December 30 2021.}}

%
%

\markboth{}%
{Yan \MakeLowercase{\textit{et al.}}: Semantics-Guided Contrastive Network for Zero-Shot Object Detection}
%



\IEEEtitleabstractindextext{%
\begin{abstract}
Zero-shot object detection (ZSD), the task that extends conventional detection models to detecting objects from unseen categories, has emerged as a new challenge in computer vision. {Most existing approaches tackle the ZSD task with a strict mapping-transfer strategy that may lead to suboptimal ZSD results: 
1) the learning process of these models neglects the available semantic information on unseen classes, which can easily bias towards the seen categories; 
2) the original visual feature space is not well-structured for the ZSD task due to the lack of discriminative information.} To address these issues, we develop a novel Semantics-Guided Contrastive Network for ZSD, named ContrastZSD, a detection framework that first brings contrastive learning mechanism into the realm of zero-shot detection. 
Particularly, ContrastZSD incorporates two semantics-guided contrastive learning subnets that contrast between region-category and region-region pairs respectively. The pairwise contrastive tasks take advantage of supervision signals derived from both the ground truth label and class similarity information. By performing supervised contrastive learning over those explicit semantic supervision, the model can learn more knowledge about unseen categories to avoid the bias problem to seen concepts, while optimizing the visual data structure to be more discriminative for better visual-semantic alignment. Extensive experiments are conducted on two popular benchmarks for ZSD, \emph{i.e.}, PASCAL VOC and MS COCO. Results show that our method outperforms the previous state-of-the-art on both ZSD and generalized ZSD tasks.
\end{abstract}

}

\maketitle

\IEEEdisplaynontitleabstractindextext

%
\IEEEpeerreviewmaketitle

\IEEEraisesectionheading{\section{Introduction}\label{sec:introduction}}

%
%
%
%
\IEEEPARstart{B}{ecause} of its importance to image understanding and analysis,
object detection has received increasing attention in recent years \cite{cai2018cascade,zhu2018attention,dai2016r,yang2021multiple}. 
With the impressive development of deep learning, a surge of novel detection models built upon deep Convolutional Neural Networks (CNNs) have been developed in recent years, pushing the detection performance forward remarkably \cite{pang2019libra,he2017mask,tian2019fcos,carion2020end,ZhouCSSYN20}.
The most state-of-the-art object detection models follow a region proposal based paradigm \cite{cai2018cascade,zhang2020dynamic,wu2020rethinking,LuLNCZ21}, which detect objects by 1) first generating region proposals as candidates that might have objects within them, and 2) then performing bounding box regression and classification simultaneously on each proposal. Despite their efficacy, the detection performance of these methods purely relies on the discriminative capabilities of region features, which often depends on sufficient training data with complete annotations for each category. However, labeling for object detection, which requires a pair of a class label and a bounding box location for each object within each image, is both prohibitively costly and labor-intensive. 
Furthermore, even if all the data samples can be well annotated, we still face the problem of data scarcity, due to the fact that novel categories (\emph{e.g.}, rare animals) are constantly emerging in practical scenarios \cite{annadani2018preserving}. In such a scenario, the traditional object detection models often become infeasible because scarce or even no visual data from those novel categories is available for model training.
The above mentioned issues, namely the burden of manual labelling and the problem of data scarcity, lead us to investigate the detection task with additional source of complexity, \emph{i.e.}, zero-shot object detection (ZSD). 

Recently, preliminary efforts have been put into the study of zero-shot object detection (ZSD) \cite{bansal2018zero,rahman2018zero,yan2020semantics,li2019zero,demirel2018zero,li2021attribute}. Most of these methods are based on a strict mapping-transfer strategy that tackles the ZSD task with a two-stage pipeline. Specifically, at the training stage, a mapping function is learned to project the visual features and semantic vectors from seen classes to a joint embedding space.
In the embedding space, the visual features can be compared directly with class embeddings using a compatibility function, which requires the score for the correct class is higher than that for all the incorrect classes.
At the testing stage, the mapping function learned on labeled visual data from seen classes is directly applied to project the visual features and semantic representations of the unseen classes into the joint embedding space, followed by nearest neighbor (NN) search for unseen class label prediction. 

\begin{figure} 	
	\centering 		
	\subfigure[]{\label{1a}		
		\includegraphics[width=1\linewidth]{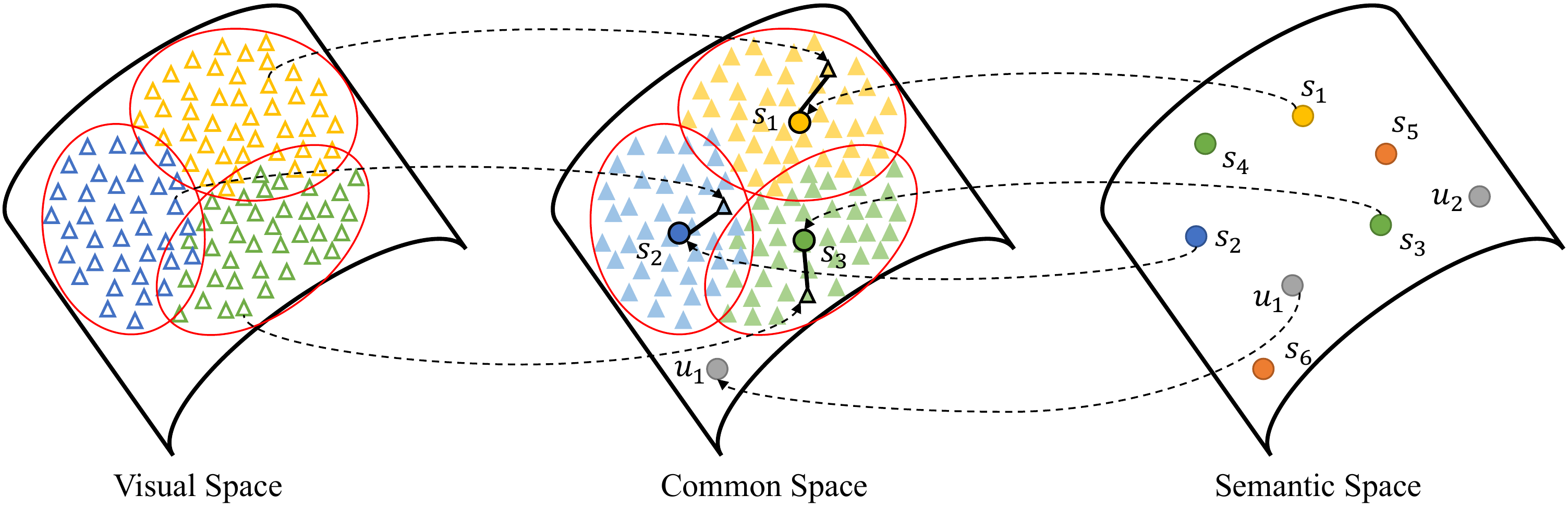}}
	\subfigure[]{\label{1b}				
		\includegraphics[width=1\linewidth]{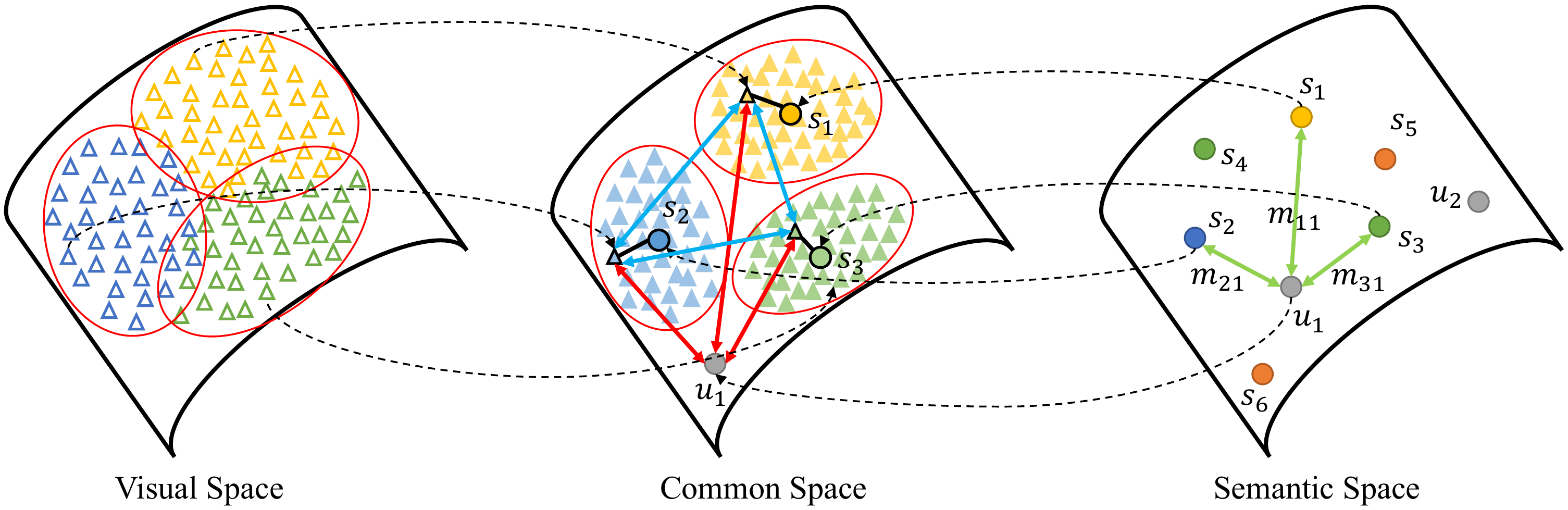}}
	\caption{Illustration of (a) the conventional embedding function based ZSD methods that rely on node-to-node projection (black dotted arrows), where $s_i$ and $u_j$ refer to seen and unseen classes respectively, and (b) the proposed ContrastZSD improves the projection with different contrastive learning strategies (red and blue arrows) under the guidance of class labels (\emph{e.g.}, $s_1,s_2,s_3$) and class similarities (\emph{e.g.}, $m_{11},m_{21},m_{31}$) for better visual-semantic alignment.}
	\label{example}	
\end{figure}

{However, since the learning process of these models merely relies on the visual data and class embeddings from seen categories, the output nodes corresponding to unseen classes are always inactive during learning.
As a result, the learned model can easily bias towards the seen categories, leading to limited transferable ability of the model \cite{fu2015transductive,ye2021alleviating,ZhangCLLWGH20}.}
Especially in the more challenging generalized zero-shot object detection (GZSD) setting, where the test samples may come from either seen or unseen classes, the bias problem would degrade the performance significantly.
The objects from unseen categories tend to be recognized as seen class objects at the test stage. 
{Additionally, the discriminative ability of visual features, which has been ignored by most previous methods, is proved to be beneficial for enhancing the ZSD/GZSD performance \cite{xie2020region}.
We have noted that the conventional mapping-transfer based ZSD methods are directly performed on the visual features extracted from pre-trained backbone networks. However, those high-dimensional visual features are far from the semantic information and thus are lack of discriminative ability. Thus, directly aligning the original visual features with the semantic information is suboptimal for ZSD. 
Taking both the model's bias problem and indistinctive visual space into consideration, a simple node-to-node projection across different spaces, as shown in Fig. \ref{1a}, may not align the visual features and semantic embeddings well.}

In this paper, we develop a semantics-guided contrastive network, namely ContrastZSD, {that seeks to optimize the visual feature distribution and simultaneously alleviate the bias problem for improved zero-shot detection. 
Particularly, instead of adopting the commonly-used mapping-transfer strategy, we develop a mapping-contrastive strategy for the proposed ContrastZSD model.}
Equipped with the similar region feature encoding network as Faster R-CNN, ContrastZSD first extracts global feature maps from the input images via CNN backbone, then produces region proposals in an objectiveness manner using the region proposal network (RPN).  
Subsequently, both the region features and semantic embeddings are mapped to a joint embedding space for visual-semantic alignment.
Unlike most existing works on ZSD that learn projection function from visual to semantic space, a common intermediate embedding space is learned in ContrastZSD, making it possible to adjust the data structures of both semantic vectors and visual features. 
As illustrated in Fig. \ref{1b}, when mapping the region features and semantic embeddings to the common space, ContrastZSD incorporates two semantics-guided contrastive learning subnets for better visual-semantic matching: 1) Region-category contrastive learning (RCCL) subnet, which is the key component that endows our model with the ability of detecting unseen objects. {This subnet employs a two-path learning mechanism to contrast seen region proposals with both seen and unseen class embeddings. Both the unseen class embeddings and unseen class similarity information can be effectively utilized in the training process, thereby alleviating the bias problem to seen categories}, and 2) Region-region contrastive learning (RRCL) subnet, which regulates the visual feature distribution to be more discriminative by resorting to the class label information.
{Specifically, this subnet contrasts between different region proposals in order to distinguish positive samples from a large number of negative samples, resulting in higher intra-class compactness and inter-class separability of the visual structure.}
To optimize the proposed deep network, we design a novel multi-task loss that includes both the classification, bounding box regression and contrastive loss. 
Our main contributions are summarized below:
\begin{itemize}
    \item We develop a novel semantics-guided contrastive network for ZSD, {underpinned by a new mapping-contrastive strategy superior to the conventional mapping-transfer strategy}. To the best of our knowledge, this is the first work that applies contrastive learning mechanism for ZSD.
    \item The proposed deep model incorporates both region-category and region-region contrastive learning to better align the region features and the corresponding semantic descriptions.
    \item In order to perform contrastive learning in a supervised paradigm, we design novel contrastive losses supervised by explicit semantic information to guarantee both the discriminative and transferable property of the proposed ZSD model.
    \item We evaluate the proposed model on two popular object detection datasets, \emph{i.e.}, PASCAL VOC and MS COCO. The experimental results show that our method performs favorably against the state-of-the-art approaches over both the ZSD and GZSD task. 
\end{itemize}

 
\section{Related Works}
In this section, we briefly review the related works on the fields relevant to our study: object detection, zero-shot learning, zero-shot object detection and supervised contrastive learning.

\noindent\textbf{Object detection.}
As one of the most important tasks in computer vision, object detection has received considerable attention and experienced significant development in the past decade. Modern object detection models can be roughly categorized into two groups. 
One follows the conventional two-stage detection pipeline, namely region proposal based methods. They first generate all possible regions of interest, then pass the region proposals to the down-stream task-specific layers for classification and box regression.
The region proposal based methods mainly include Faster R-CNN \cite{ren2015faster}, R-FCN \cite{dai2016r}, FPN \cite{lin2017feature}, SPP-net \cite{he2015spatial} and Mask R-CNN \cite{he2017mask}.
The other popular group, referred as one-stage detection models, adopts a single-step regression pipeline to map straightly from image pixels to bounding box coordinates and class probabilities.
The most representative methods in this group include SSD \cite{liu2016ssd}, YOLO \cite{redmon2017yolo9000}, FCOS \cite{tian2019fcos} and RetinaNet \cite{lin2017focal}.
Despite the empirical success on object categories with sufficient training data, these methods are unable to deal with the detection problem on novel concepts without training samples. 
In general, one-stage methods with a simple single-step pipeline enjoy reduced time expense, but typically achieve lower accuracy rates than region proposal based methods. Thus, we here focus on tackling the ZSD problem with region proposal based detection models due to their high performance.

\noindent\textbf{Zero-Shot Learning.}
The previous research literature on zero-shot learning exhibits great diversity, such as learning independent attribute classifiers \cite{lampert2013attribute,ChangYLZH16,ChangYHXY15,Zhang0LCZ18,ChangYYX16}, learning embedding functions \cite{lei2015predicting,yang2014unified,frome2013devise,socher2013zero} and generative adversarial networks based methods \cite{xian2018feature,felix2018multi}. In this section, we focus on the embedding based methods that are the most relevant to ours.
The key idea of those methods is to learn an embedding function that maps the semantic vectors and visual features into an embedding space, where the visual features and semantic vectors can be compared directly. 
Embedding based ZSL methods differ in what embedding space is employed, which can be broadly divided into three types: learning a common embedding space for visual space and semantic space \cite{lei2015predicting,yang2014unified}, learning an embedding from visual space to semantic space \cite{frome2013devise,socher2013zero,xian2016latent,akata2015evaluation}, and learning an embedding from semantic space to visual space \cite{shigeto2015ridge}.
Among those embedding strategies, the common intermediate embedding space makes it possible to adjust data structures of both semantic vectors and visual features \cite{wang2019visual}. 
Thus, the common intermediate space is employed in our work.

\noindent\textbf{Zero-Shot Object detection.}
ZSD is a recently introduced task in \cite{rahman2018zero} and still remains under-explored in the computer vision literature.
Most existing methods on ZSD focus on learning an embedding function from visual to semantic space \cite{bansal2018zero,rahman2018zero,yan2020semantics,li2019zero,demirel2018zero}.
For example, Rahman \emph{et al.} \cite{rahman2018zero} developed a Faster R-CNN based semantic alignment network with a novel semantic clustering loss for ZSD.
Considering the ambiguous nature of background class in ZSD, Bansal \emph{et al.} \cite{bansal2018zero} designed several background-aware detectors to address the confusion between unseen and background objects using external annotations.
Demirel \emph{et al.} \cite{demirel2018zero} developed a hybrid region embedding model that joins a convex combination of semantic embeddings with an object detection framework.
Apart from those visual-to-semantic mapping methods, there also exist some methods that learn a common space between visual space and semantic space \cite{li2019zero} or learn an embedding from semantic to visual space \cite{zhang2020zero,gupta2020multi}. 
{Despite their efficacy, all of these methods fail to consider the two key factors that impair the ZSD performance, \emph{i.e.}, the model's bias problem to seen classes and the indistinctive visual space.}
To alleviate these issues, the proposed model goes further to bring contrative learning mechanism into the realm of ZSD, allowing for further improvement of ZSD performance. 

\noindent\textbf{Contrastive Learning.}
Contrastive learning, which can be considered as learning by comparing, has achieved significant advancement in self-supervised representation learning \cite{chen2020simple,he2020momentum,tian2020makes,liu2021self}. Recently, a trend has emerged of leveraging contrastive learning to facilitate self-supervised computer vision tasks \cite{xie2021detco,park2020contrastive,qian2021spatiotemporal}. First, a number of positive/negative samples are usually created for each anchor image through data augmentation. 
Then, contrastive learning is performed between positive and negative pairs of images against each other, with the objective of pulling the representation of ``similar'' samples together and pushing that of ``dissimilar'' samples further away in the embedding space.
However, contrastive learning used in those self-supervised algorithms fails to consider the high-level class semantics since they assign only the augmented view for each image. For this issue, a few approaches have been proposed to leverage human-annoted labels, which has been shown to be more robust to corruption.
For example, Khosla \emph{et al.} \cite{khosla2020supervised} directly used class labels to define similarity, where samples from the same class are positive and samples from different classes are negative samples. 
Majumder \emph{et al.} \cite{majumder2021revisiting} devised few-shot learning with Instance discrimination based contrastive learning in a supervised setup.
Inspired by the success of these methods, we first introduce contrastive learning mechanism to ZSD, and develop two contrastive learning subnets that utilize high-level semantic information as additional supervision signals.
\section{The Proposed Method}
This section begins with the problem setting of ZSD in Section 3.1. Section 3.2 describes the overall model framework. Next, we introduce the semantics-guided contrastive learning subnets in Section 3.3. Finally, Section 3.4 discusses the training and inference details of the proposed network.
\subsection{Problem Formulation}
\textbf{Notations.} In the framework of ContrastZSD, we denote the set of all classes as $\mathcal{Y}=\mathcal{Y}^f\cup\{y_0\}$, where $\mathcal{Y}^f$ denotes the set of all foreground classes and $y_0$ refers to the background class. More specifically, $\mathcal{Y}^f$ can be decomposed into two disjoint subsets, \emph{i.e.}, $\mathcal{Y}^f=\mathcal{Y}^s\cup\mathcal{Y}^u$ ($\mathcal{Y}^s\cap\mathcal{Y}^u=\varnothing$), where $\mathcal{Y}^s=\{y_1, y_2, \cdots, y_{n_s}\}$ and $\mathcal{Y}^u=\{y_{n_s+1}, y_{n_s+2}, \cdots, y_{n_s+n_u}\}$ denote the set of seen and unseen classes respectively.
Given all the classes defined above, the whole label space turns to be $\mathcal{Y}=\{y_0, y_1, y_2, \cdots, y_{n_s+n_u}\}$ with the cardinality being $n_c=n_s+n_u+1$.
Inspired by previous works on ZSL, each foreground class in $\mathcal{Y}$ can be represented by a $d_c$-dimensional semantic embedding generated in an unsupervised manner from external linguistic sources, such as Word2Vec\cite{mikolov2013distributed} or Glove \cite{pennington2014glove}.
Considering the ambiguous nature of the background class, it's unfeasible to learn a fixed class embedding from off-the-shelf linguistic sources for it.
In order to reduce the confusion between the background and unseen objects, we follow the Background Learnable RPN developed in \cite{zheng2020background} to learn a discriminative semantic vector $a_0$ for the background class $y_0$.
We denote $\mathbf{A}=[\mathbf{a}_{y_0},\mathbf{a}_{y_1},\mathbf{a}_{y_2},\dots,\mathbf{a}_{y_{n_s+n_u}}]\in\mathbb{R}^{n_c\times d_c}$ as the matrix that collects the semantic embeddings of all the categories; here, $\mathbf{a}_{y_i}$ refers to the label embedding of class $y_i$ in $\mathcal{Y}$.

\noindent\textbf{Category Similarity.} 
{To allow for semantic relation guided knowledge transfer, we further introduce a class similarity matrix $\mathbf{S}=[\mathbf{s}_0,\mathbf{s}_1,\dots,\mathbf{s}_{n_s+n_u}]\in\mathbb{R}^{n_c\times n_u}$, where $\mathbf{s}_i=\{s_{ij}\}\in\mathbb{R}^{n_u}$ is the $i$-th row vector of $\mathbf{S}$ that characterizes the unseen distribution of the $i$-th class in $\mathcal{Y}$; $s_{ij}$ is the $j$-th element of $\mathbf{s}_i$.
Specifically, for each seen category $y_i\in\mathcal{Y}^s$, its semantic relation to unseen class $y_j\in\mathcal{Y}^u$ is obtained by computing the cosine similarity of their corresponding semantic embeddings $\mathbf{a}_{y_i}$ and $\mathbf{a}_{y_j}$, \emph{i.e.}, $s_{ij}=\frac{\mathbf{a}_{y_i}\cdot\mathbf{a}_{y_j}}{\|\mathbf{a}_{y_i}\|_2\|\mathbf{a}_{y_j}\|_2}$. 
All the similarity values corresponding to a seen category are squashed by Softmax to acquire unseen probabilities.
For each unseen class $y_i\in\mathcal{Y}^u$, its unseen distribution is set as a one-hot vector $\mathbf{s}_i\in\{0,1\}^{n_u}$ with $s_{ij}=0(j\neq i)$ and $s_{ii}=1$, while the unseen distribution of background class $y_0$ is set as zero vector, \emph{i.e.}, $\mathbf{s}_{0}=\mathbf{0}\in\mathbb{R}^{n_u}$.}

\noindent\textbf{Task Definition.} Assuming an image set $\mathcal{X}$ that includes $n$ images about $n_s+n_u$ object categories is provided for detection.
Each image in $\mathcal{X}$ consists of several objects with boxes $\mathcal{B}=\{b_i\}_{i=1}^{n_r}$ and ground truth labels $\{c_i\}_{i=1}^{n_r}$, where $n_r$ is the number of boxes; $b_i=(x_i, y_i,w_i,h_i)$ is the $i$-th object box with $c_i$ being the ground truth label. 
More specifically, $\mathcal{X}$ is composed of two subsets, \emph{i.e.}, $\{\mathcal{X}^{tr}, \mathcal{X}^{te}\}$, where $\mathcal{X}^{tr}$ and $\mathcal{X}^{te}$ correspond to the training and testing image set respectively.
The training image set $\mathcal{X}^{tr}=\{x_1, x_2, \cdots, x_{n_{tr}}\}$ collects $n_{tr}$ labeled visual data that contain only objects from seen categories $\mathcal{Y}^s$, while the images in testing set $\mathcal{X}^{te}=\{x_{n_{tr}+1}, x_{n_{tr}+2}, \cdots, x_{n_{tr}+n_{te}}\}$ contain objects belonging to testing categories $\mathcal{Y}^{te}$. Notably, the definition of testing category set $\mathcal{Y}^{te}$ depends on the task settings, where $\mathcal{Y}^{te}=\mathcal{Y}^{u}$ for ZSD and $\mathcal{Y}^{te}=\mathcal{Y}^{s}\cup\mathcal{Y}^{u}$ for GZSD respectively.
Under the guidance of the common semantics among seen and unseen classes, \emph{i.e.}, $\mathbf{A}$ and $\mathbf{S}$, our ContrastZSD model is trained on the seen object annotations over the training set $\mathcal{X}^{tr}$, with the objective of generalizing to the detection of unseen objects in $\mathcal{X}^{te}$. Specifically, for each test image in $\mathcal{X}^{te}$, the task of ZSD or GZSD is to recognize all the foreground objects in the image that belong to the testing categories $\mathcal{Y}^{te}$ and simultaneously localize their bounding box coordinates. 

\subsection{Model Architecture}
\begin{figure*}[t] 	
	\centering 		
	\includegraphics[width=1\linewidth]{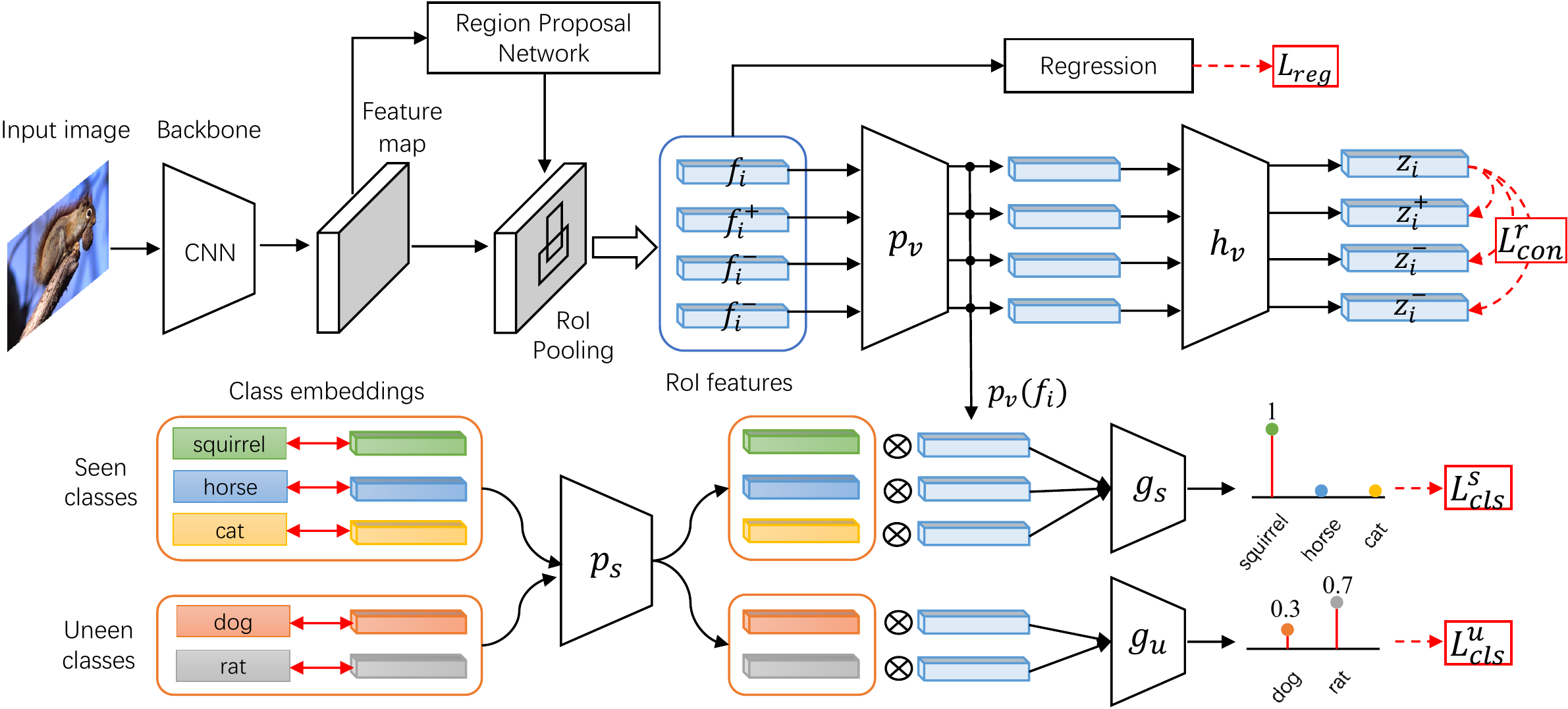}		
	\caption{Illustration of the proposed ContrastZSD framework. First, the region feature encoding network takes raw images as input to produce region proposals and object-level visual features, followed by box regression branch for coordinate offset prediction. Then, the region features and class embeddings are mapped to a joint embedding space by using embedding function $p_v$ and $p_s$ respectively. Next, region-region contrastive learning is conducted with projection head $h_v$ to optimize the visual feature distribution with higher discriminability. Simultaneously, region-category contrastive learning is performed over the mapped region features and class embeddings in order to improve the transferable ability of the model, where $g_s$ and $g_u$ are projection heads developed for seen and unseen classes respectively; $\otimes$ refers to element-wise product.}
	\label{framework}	
\end{figure*}
In this section, we systematically introduce the overall architecture of the proposed ContrastZSD, including region feature encoding subnet, bounding box regression subnet and visual-semantic alignment subnet.
The overall framework of ContrastZSD is shown in Fig. \ref{framework}. Here, we elaborate each subnet in detail.

\subsubsection{Region Feature Encoding Subnet}
\textbf{CNN Backbone.} Given an arbitrary image as input, the CNN backbone network produces intermediate convolutional activations as image-level feature map.
In our work, the basic architecture of the CNN backbone is ResNet composed of four convolutional blocks (conv1 to conv4) \cite{he2016deep}, where the output of conv4 module is regarded as the global feature map of input image.
Taking a RGB image with dimension $\mathbb{R}^{3\times H\times W}$ as input, the output of the CNN backbone network is a feature map with dimension $\mathbb{R}^{1024\times\frac{H}{16}\times\frac{W}{16}}$, where $H$ and $W$ denote the height and width of the input image respectively with the output channel being 1024.
Subsequently, the image-level feature maps will be fed into the region proposal network to generate regions of interest (RoIs) and object-level features.

\noindent\textbf{Region Proposal Network (RPN).} Taking the image's global feature map as input, the RPN first generates $k$ anchor boxes at each sliding window location of the feature map. Specifically, we set $k$ as 9 in our work, and thus the total number of anchor boxes is $9\times\frac{H}{16}\times\frac{W}{16}$.
Then, all the anchor boxes are fed into two modules: (1) the classification module scores each proposal as either an object (positive anchor) or background (negative anchor); (2) the box regression module predicts the coordinate offsets for each region proposal. Next, it ranks the positive anchors at each sliding window location, then generates a set of candidate object proposals (RoIs) after applying the predicted offsets, denoted as $\mathcal{R}=\{r_i\}_{i=1}^{n_r}$. Since the top ranking candidate proposals can be of variable sizes, a RoI-pooling layer is further applied to project the visual features of varying-sized proposals to fixed-dimensional representations. 
We denote $\mathbf{F}=[\mathbf{f}_1,\mathbf{f}_2,\dots,\mathbf{f}_{n_r}]\in\mathbb{R}^{n_r\times d_r}$ as the feature representation matrix corresponding to $\mathcal{R}$, where $\mathbf{f}_i$ refers to the visual feature of proposal $r_i$ with $d_r$ being the feature dimension.
The subsequent parts of the model aim to construct an effective zero-shot detector using the extracted region features and class semantics.

\subsubsection{Visual-Semantic Alignment Subnet}
\textbf{Mapping-Transfer Strategy.}
Most existing methods on ZSD are based on a strict mapping-transfer strategy, where the mapping function is learned on seen classes then transferred directly to unseen classes.
The mapping function connects the visual features and auxiliary semantic descriptions by projecting them into a joint embedding space, such that they can be compared directly.
The space spanned by semantic embeddings is often chosen as the embedding space in previous works \cite{bansal2018zero, rahman2018zero,li2019zero}.
After projecting the extracted visual region features to semantic space, a compatibility function $\mathcal{S}(\mathbf{W}_p^\top\mathbf{f}_i,\mathbf{a}_{y_j})$ is employed to measure the matching degree between the projected proposal $r_i$ and class $y_j$, where $\mathbf{W}_p\in\mathbb{R}^{d_r\times d_c}$ is the trainable weight matrix of the projection layer.
The mapping function is usually trained by standard cross entropy loss or max-margin loss to facilitate the separation between ground truth class and the rest classes \cite{li2018deep,demirel2018zero}.

At the testing stage, the model trained on seen classes is directly generalized to the detection of unseen objects. Given a test sample $r$, the label prediction is performed by simply selecting the most matching unseen category in the embedding space, \emph{i.e.},
\begin{align}
y^*=\arg\max_{y_k\in \mathcal{Y}^u}\mathcal{S}(\mathbf{W}_p^\top \mathbf{f},\mathbf{a}_{y_k}),
\end{align}
where $\mathbf{f}\in\mathbb{R}^{d_r}$ denotes the visual feature of test sample $r$.

The key to these methods is learning an exact projection by tightly mapping labeled visual data to their corresponding seen classes.
{We conjecture that such a strict projection constraint can easily bias the learning process towards seen categories, and thus sacrifice the model's generalization ability to unseen classes. Moreover, the visual-semantic projection tends to be suboptimal for the ZSD task, since it is performed on the original feature space that is usually not well-structured.} To this end, we develop a novel mapping-contrastive strategy to improve the conventional mapping-transfer strategy through semantics-guided contrastive learning.

\noindent\textbf{Mapping-Contrastive Strategy.}
Considering the huge gap between visual and semantic spaces, we treat the common intermediate space as the embedding space to optimize the data structures of both visual features and semantic descriptions.
First, we employ two mapping functions, \emph{i.e.}, $p_v(\cdot)$ and $p_s(\cdot)$, to embed the visual features and semantic descriptions into the joint embedding space respectively:
\begin{align}
    p_v(\mathbf{f}_i)&=\delta(\mathbf{W}_v\mathbf{f}_i+\mathbf{b}_v),\\
    p_s(\mathbf{a}_{y_j})&=\delta(\mathbf{W}_s\mathbf{a}_{y_j}+\mathbf{b}_s),
\end{align}
where $p_v(\cdot)$ and $p_s(\cdot)$ are implemented as Multi-Layer Perceptron (MLP) network with $\mathbf{W}_v$, $\mathbf{W}_s$ being the trainable weight matrix and $\mathbf{b}_v$, $\mathbf{b}_s$ denoting the bias; $\delta$ refers to the nonlinear activation. 
Subsequently, the mapped visual features $p_v(\mathbf{f}_i)$ and semantic descriptions $p_s(\mathbf{a}_{y_j})$ in the joint embedding space are fed into two semantics-guided contrastive branches, \emph{i.e.}, region-category contrastive learning (RCCL) and region-region contrastive learning (RRCL), for better visual-semantic alignment.

Different from the mapping-transfer strategy that employs a fixed compatibility function $\mathcal{S}(\cdot)$, the RCCL subnet in our model automatically judges how well the object is consistent with a specific class through contrastive learning. To avoid biasing to the seen classes, it contrasts seen region features with unseen class embeddings guided by the similarity information between seen and unseen classes. In this way, the proposed model can explicitly transfer knowledge from seen classes to unseen categories during the training phase, leading to improved generalization ability.

Additionally, the RRCL subnet regulates the visual data distribution in the joint embedding space by contrasting between different region proposals. Under the guidance of class label information, samples belonging to the same class are pulled together in embedding space, while simultaneously pushing apart samples from different classes. As a result, our model can produce more discriminative region features with high intra-class compactness and large inter-class margin, with a boost to visual-semantic alignment as byproduct.
We will introduce the details of each contrastive learning branch in Section 3.3.

\subsubsection{Bounding Box Regression Subnet}
Unlike the image classification task containing only classification results, object detection also predicts object locations, which is performed by adding suitable offsets to the generated regions in order to align them with the ground truth coordinates. 
In this work, we employ the standard bounding box regression branch in Faster R-CNN, considering that it is class-agnostic and transferable enough.

Given a detected bounding box $b_i=(x_i^o,y_i^o,w_i^o,h_i^o)$ and its ground truth box coordinates $(x_i^t,y_i^t,w_i^t,h_i^t)$, the regressor is configured to learn scale-invariant transformation between two centers and log-scale transformation between widths and heights. The ground truth offsets $t_i^{*}=(t_{ix}^{*},t_{iy}^{*},t_{iw}^{*},t_{ih}^{*})$ can be computed as: 
\begin{align}
\begin{split}
&t_{ix}^{*}=\frac{|x_i^t-x_i^o|}{x_i^o},\ \ \ \  t_{iy}^{*}=\frac{|y_i^t-y_i^o|}{y_i^o},\\
&t_{iw}^{*}=\log\frac{w_i^t}{w_i^o},\ \ \ \ t_{ih}^{*}=\log\frac{h_i^t}{h_i^o}. 
\end{split}
\end{align}
At the same time, the box regression layer takes the visual feature $\mathbf{f}_i\in\mathbb{R}^{d_r}$ as input to predict the coordinate offsets as $t_i=(t_{ix}, t_{iy},t_{iw},t_{ih})$.
Subsequently, we minimize the regression loss for all the $n_r$ region proposals, \emph{i.e.},
\begin{align}
\label{regression}
L_{reg}=\sum_{i=1}^{n_r}\sum_{j\in\{x,y,w,h\}}\text{smooth}_{\ell_1}(t_{ij}-t_{ij}^*),
\end{align}
where $\text{smooth}_{\ell_1}(\cdot)$ denotes the same smooth $\ell_1$ loss used in Faster R-CNN that tweaks the predicted region coordinates to the corresponding target bounding box.

\subsection{Semantics-Guided Contrastive Learning}
The ZSD task aims to train an effective zero-shot detector that is both ``discriminative'' enough to discriminate between seen classes and ``transferable'' well to unseen classes. 
Accordingly, in this section, we introduce two semantics-guided contrastive learning branches to guarantee both the discriminative and transferable ability of the proposed ContrastZSD model.

\subsubsection{Region-Category Contrastive Learning}

\noindent\textbf{Visual-Semantic Consistency.}
During the training stage, only the visual features from seen categories are provided, while the semantic embeddings corresponding to both seen and unseen classes are available to access. In order to enable explicit knowledge transfer from seen to unseen domain, we propose to contrast the visual features of seen objects with both seen and unseen classes to distinguish whether they are consistent or not. Recall that we have encoded the visual features of seen region proposals and all the class embeddings into the joint embedding space in Section 3.2. For each region-category pair $(\mathbf{f}_i, \mathbf{a}_{y_j})$ encoded as $\langle p_v(\mathbf{f}_i),p_s(\mathbf{a}_{y_j})\rangle$, we first fuse their information and then distinguish how consistent the fusion is. {The consistency value between $p_v(\mathbf{f}_i)$ and $p_s(\mathbf{a}_{y_j})$ can be computed as
\begin{align}
\label{consistency}
\begin{split}
o(\mathbf{f}_i, \mathbf{a}_{y_j})=\left\{
\begin{array}{ll}
\sigma(g_s(p_v(\mathbf{f}_i)\otimes p_s(\mathbf{a}_{y_j}))), & \hbox{$y_j\in \mathcal{Y}^s\cup\{y_0\}$}, \\
\sigma(g_u(p_v(\mathbf{f}_i)\otimes p_s(\mathbf{a}_{y_j}))), & \hbox{$y_j\in \mathcal{Y}^u$},
\end{array}
\right.
\end{split}
\end{align}
where $\otimes$ refers to the element-wise product operation for visual-semantic information fusion; $g_s(\cdot)$ and $g_u(\cdot)$ are two independent projection heads developed for seen and unseen classes respectively to enable two-path learning, both of which are implemented as MLP network with $\sigma$ being the nonlinear activation. }

\noindent\textbf{Consistency Based Classification Branch.}
Given $n_r$ region proposals, we first operate Eq. (\ref{consistency}) over all the classes to predict the consistency scores in matrix form, {denoted as $\mathbf{O}^s=\{o_{ij}^s\}\in\mathbb{R}^{n_r\times(n_s+1)}$ and $\mathbf{O}^u=\{o_{ij}^u\}\in\mathbb{R}^{n_r\times n_u}$ for seen and unseen classes respectively.}
Then, we utilize not only the ground truth labels $\{c_i\}_{i=1}^{n}$ but also the class similarity matrix $\mathbf{S}$ as supervision signals, and derive the full classification loss as
\begin{align}
\label{rccl1}
L_{cls}=L_{cls}^s+\lambda L_{cls}^u,
\end{align}
where $L_{cls}^s$ and $L_{cls}^u$ refer to the classification loss with respect to seen and unseen classes respectively; $\lambda$ is a trade-off parameter.
More specifically, $L_{cls}^s$ is designed to endow the model with the discriminative ability to distinguish different seen classes, which can be formulated as the general cross-entropy loss, \emph{i.e.},
\begin{align}
\label{rccl2}
L_{cls}^s=-\sum_{i=1}^{n_r}c_i\log o_{ic_i}^s,
\end{align}
where $c_i$ refers to the ground truth label index of the $i$-th proposal. 
To enable explicit knowledge transfer, we further take advantage of the class similarity information in $\mathbf{S}$ as additional supervision signals. {For each training sample, the predicted unseen class probability in $\mathbf{O}^u$ should be consistent with the corresponding unseen class similarity pre-defined in $\mathbf{S}$.}
Thus, $L_{cls}^u$, the second item in Eq. (\ref{rccl1}), turns to be the following binary cross-entropy loss:
\begin{align}
\label{rccl3}
L_{cls}^u=-\sum_{i=1}^{n_r}\sum_{j=1}^{n_u}s_{c_ij}\log o_{ij}^u+(1-s_{c_ij})\log(1- o_{ij}^u).
\end{align}
{By minimizing $L_{cls}^u$, we can learn unseen class predictors using training proposals from a group of related seen classes as their pseudo-instances, thereby alleviating the bias problem caused by the lacking of training data.}
In this way, the model's transferable capability to the unseen domain can be strengthened without disturbing the seen object detection optimized by $L_{cls}^s$.

\subsubsection{Region-Region Contrastive Learning}
The key to region-category contrastive learning lies in that the embedded semantic vector of one class should try to be consistent with every visual instance features from this class. However, the distribution of instances in the original visual space tends to be indistinctive and thus is suboptimal for zero-shot detection. The case can be even worse for the object detection task since the top ranking proposals may only cover parts of objects instead of the whole objects. Motivated by this, we propose to contrast between different region proposals with the help of semantic label information, with the objective of optimizing the visual data structure to be more discriminative in the joint embedding space.
Given $n_r$ region proposals generated from the same batch of images, we first map their features $\{p_v(\mathbf{f}_i)\}_{i=1}^{n_r}$ to new representations $\{\mathbf{z}_i\}_{i=1}^{n_r}$ using an embedding network $h_v(\cdot)$:
\begin{align}
\label{eq4}
\mathbf{z}_i=h_v(p_v(\mathbf{f}_i))=\theta(\mathbf{W}_{h_v} p_v(\mathbf{f}_i)+\mathbf{b}_{h_v}),
\end{align}
where $h_v(\cdot)$ is a MLP network with weight matrix $\mathbf{W}_{h_v}$ and bias $\mathbf{b}_{h_v}$; $\theta$ refers to nonlinear activation.

Unlike the conventional self-supervised contrastive learning that focuses only on instance discrimination, we aim to achieve class discrimination by effectively leveraging the label information.
For each region proposal $r$, we treat the proposals from the same class with $r$ as positive samples, and all the other proposals generated from the same batch of images as negative samples.
Taking the $i$-th region proposal encoded as $\mathbf{z}_i$ as an example, we assume that there are $N_i^p$ positive proposals $\{\mathbf{z}_1^+, \mathbf{z}_2^+, \cdots, \mathbf{z}_{N_i^p}^+\}$ and $N_i^n$ negative samples $\{\mathbf{z}_1^-, \mathbf{z}_2^-, \cdots, \mathbf{z}_{N_i^n}^-\}$. Each positive sample $\mathbf{z}^+$ shares the same label with $\mathbf{z}_i$, while the class label of $\mathbf{z}^-$ is different from $\mathbf{z}_i$.
The region-region contrastive loss used for a pair of bounding boxes $(\mathbf{z}_i, \mathbf{z}_{j}^+)$ takes the following form,
\begin{align}
\label{eq9}
\ell_{con}^r(\mathbf{z}_i, \mathbf{z}_{j}^+)=-\log\frac{\exp(\frac{\mathbf{z}_i\cdot \mathbf{z}_j^+}{\tau})}{\sum_{k=1}^{N_i^p}\exp(\frac{\mathbf{z}_i\cdot \mathbf{z}_k^+}{\tau})+\sum_{k=1}^{N_i^n}\exp(\frac{\mathbf{z}_i\cdot \mathbf{z}_k^-}{\tau})},
\end{align}
where $\tau$ is the temperature parameter set as 0.1 by default as in \cite{khosla2020supervised}. Thus, the total contrastive loss $L_{con}^r$ for $n_r$ region proposals can be formulated as
\begin{align}
\label{eq10}
L_{con}^r=\sum_{i=1}^{n_r}\frac{1}{N_i^p}\sum_{j=1}^{N_i^p}\ell_{con}^r(\mathbf{z}_i, \mathbf{z}_{j}^+).
\end{align}
Benefiting from the constraint in Eq. (\ref{eq10}), the region features from the same class is pulled closer, while the instances from different classes are pushed farther apart, resulting in a more distinguishable visual data structure.

\begin{algorithm}[t]\scriptsize
\SetAlgoLined
    \PyComment{e\_r: region feature encoding network} \\
    \PyComment{p\_v, p\_s: mapping functions for visual and semantic} \\
    \PyComment{g\_s, g\_u: projection heads for seen and unseen classes} \\
    \PyComment{h\_v: embedding network of RRCL} \\
    \PyComment{A, S: class embedding matrix, unseen distribution matrix} \\
    \PyComment{n\_s, n\_u: number of seen and unseen classes} \\
    \PyCode{for x in data\_loader:} \PyComment{minibatch of N samples} \\
    \Indp   
        \PyComment{generate region proposals (RoIs)}\\
        \PyCode{RoI, RoI\_fea, RoI\_label, RoI\_target = e\_r.forward(x)}\\
        \BlankLine
        \BlankLine
        \PyComment{predict  the  coordinate  offsets} \\
        \PyCode{offset\_p = regressor.forward(RoI\_fea)}\\
        \PyComment{box regression loss, Eqn.(\ref{regression})} \\
        \PyCode{loss\_reg = smooth\_$\ell_1$\_loss(offset\_p, RoI\_target)}\\
        \BlankLine
        \BlankLine
        \PyComment{region-region contrastive learning} \\
        \PyCode{fea\_p = p\_v.forward(RoI\_fea)} \PyComment{visual mapping} \\
        \PyCode{Z = h\_v.forward(fea\_p)} \PyComment{embedding for RRCL}\\
        \PyComment{contrastive loss, Eqn.(\ref{eq9}) and (\ref{eq10})} \\
        \PyCode{loss\_con = contras\_loss(Z, RoI\_label)}\\
        \BlankLine
        \BlankLine
        \PyComment{seen object classification} \\
        \PyCode{A\_p = p\_s.forward(A)} \PyComment{semantic mapping} \\
        \PyCode{bs = fea\_p.size(0)} \PyComment{batch size of RoIs} \\
        \PyCode{fea\_s = fea\_p.unsqueeze(0).repeat(n\_s+1, 1, 1).transpose(0, 1)} \PyComment{fea\_s dim: bs*(n\_s+1)*d\_r}\\
        \PyCode{A\_s = A\_p[:n\_s+1, :].unsqueeze(0).repeat(bs, 1, 1)}\PyComment{A\_s dim: bs*(n\_s+1)*d\_r}\\
        \PyCode{O\_s = g\_s.forward(fea\_s*A\_s)} \PyComment{contrastive values}\\
        \PyComment{classification loss for seen, Eqn.(\ref{rccl2})} \\
        \PyCode{loss\_cls\_s = CrossEntropyLoss(O\_s, RoI\_label)}\\
        \BlankLine
        \BlankLine
        \PyComment{unseen probability prediction} \\
        \PyCode{fea\_u = fea\_p.unsqueeze(0).repeat(n\_u, 1, 1).transpose(0, 1)} \PyComment{fea\_u dim: bs*n\_u*d\_r}\\
        \PyCode{A\_u = A\_p[-n\_u:, :].unsqueeze(0).repeat(bs, 1, 1)} \PyComment{A\_u dim: bs*n\_u*d\_r}\\
        \PyCode{O\_u = g\_u.forward(fea\_u*A\_u)} \PyComment{contrastive values}\\
        \PyComment{classification loss for unseen, Eqn.(\ref{rccl3})} \\
        \PyCode{loss\_cls\_u = BinaryCrossEntropyLoss(O\_u, S)}\\
        \BlankLine
        \BlankLine
        \PyComment{overall loss and SGD update} \\
        \PyCode{loss = loss\_reg+loss\_cls\_s+$\lambda$loss\_cls\_u+$\beta$loss\_con}\\
        \PyCode{optimizer.zero\_grad()}\\
        \PyCode{loss.backward()}\\
        \PyCode{optimizer.step()} \PyComment{update network params}\\
    \Indm 
\caption{Pseudo-code of the proposed ContrastZSD in Pytorch-style.}
\label{algo:your-algo}
\end{algorithm}

\subsection{Training and Inference Details}
\textbf{Training.}
Unlike previous works on ZSD that usually rely on multi-step training, we adopt an end-to-end training mechanism to jointly optimize the network parameters. We keep the bottom layers fixed to the weights pre-trained on ImageNet \cite{russakovsky2015imagenet}, and then train the RPN, bounding box regression and visual-semantic alignment network. 
More specifically, the RPN is trained with the same classification and regression loss as in Faster R-CNN. 
Notably, the RPN, which is trained on seen visual data without the exploitation of any semantic information, can generate proposals for unseen objects also, since it is designed to generate object proposal based on the objectness measure.
To optimize the proposed contrastive network, we minimize a multi-task loss specifically designed for ZSD, including both the classification, bounding box regression and contrastive losses. The overall ZSD loss takes the following form:
\begin{align}
\label{total_loss}
L_{zsd}=L_{reg}+L_{cls}^s+\lambda L_{cls}^u+\beta L_{con}^r,
\end{align}
where $\lambda$ and $\beta$ are hyper-parameters that control the trade-off between the loss terms in Eq. (\ref{total_loss}). {The pseudo-code of the proposed ContrastZSD is shown in Algorithm \ref{algo:your-algo}.}

\noindent\textbf{Inference.}
Given a test image $I^{te}\in\mathcal{X}^{te}$, we first forward $I^{te}$ into the trained region feature encoding subnet to get all the region proposals $\mathcal{R}=\{r_j\}_{j=1}^m$.
For each proposal $r_j$ with visual feature $\mathbf{f}_j$, we can generate its coordinate offsets $t_j$ by the box regressor and classification scores $\mathbf{o}_j$ by the two-path visual-semantic alignment subnet.
Specifically, the first path $g_u(\cdot)$ merely produces the unseen probability $\mathbf{o}_j^u\in\mathbb{R}^{n_u}$, while the second path $g_s(\cdot)$ takes both the seen and unseen class embeddings as input to generate the contrastive values $\mathbf{o}_j^s\in\mathbb{R}^{n_c}$ for all the classes. 
Then, $\mathbf{o}_j^s\in\mathbb{R}^{n_c}$ and $\mathbf{o}_j^u\in\mathbb{R}^{n_u}$ are fused through an extra calculation process to jointly determine the final scores, \emph{i.e.},
\begin{align}
    \mathbf{o}_j= (\mathbf{o}_j^u\mathbf{S}^\top)\otimes \mathbf{o}_j^s.
\end{align}
Finally, Non-Maximum Suppression (NMS) is applied to remove the proposals with small Intersection over Union (IoU) values and get the final detection results, where IoU is used to measure the overlap between the predicted and ground truth bounding boxes.

\section{Experiments}
\subsection{Experimental Setup}
\noindent\textbf{Datasets.}
We evaluate the proposed ContrastZSD model on two widely-used datasets
for object detection, \emph{i.e.}, PASCAL VOC 2007+2012 \cite{everingham2010pascal} and MS COCO 2014 \cite{lin2014microsoft}. 
PASCAL VOC consists of 20 common object categories for object class recognition.
More specifically, PASCAL VOC 2007 contains 2501 training images, 2510 validation images and 5011 test images. PASCAL VOC 2012 was released without test images provided, and includes 5717 training images and 5823 validation images. 
MS COCO was designed for object detection and semantic segmentation tasks. It contains 82783 training and 40504 validation images from 80 categories. 

Being zero-shot, each dataset should be split into the combination of seen/unseen subsets. For the purpose of fair comparison, we follow previous works that also target on the ZSD task to split the datasets.
For the PASCAL VOC dataset, we adopt the same setting in \cite{demirel2018zero} to split the 20 categories, where 16 classes are selected as seen and the remaining 4 are unseen classes.
In terms of the MS COCO dataset, we follow the same procedures described in \cite{bansal2018zero} to divide the dataset into two different splits: 1) 48 seen and 17 unseen classes, and 2) 65 seen and 15 unseen classes.
Based on the above seen/unseen class splits, we follow the steps in \cite{rahman2018polarity} to create the train and test set for each dataset.

\noindent\textbf{Implementation Details.}
As for the class semantic embeddings, we use the $\ell_2$ normalized 300-dim Word2Vec for MS COCO classes, which is produced by a model trained on a Wikipedia corpus in an unsupervised manner \cite{mikolov2013distributed}. For PASCAL VOC classes, we use the average of 64-dim binary per-instance attributes of all training images from aPY dataset \cite{farhadi2009describing}.
The image scale is resized to the shorter edge of 600, while keeping the original image aspect ratio.
We perform horizontal flip for augmenting the training data.
Non-Maximum Suppression (NMS) with an IoU threshold of 0.7 is employed to remove redundant bounding boxes. 
We adopt ResNet-101 \cite{he2016deep} pretrained on ImageNet \cite{russakovsky2015imagenet} as the CNN backbone. 
The mapping functions $p_v$ and $p_s$ are implemented as two fully-connected layers, taking 2048-dim region features and $d_c$-dim semantic embeddings as input respectively, then transform them to the same dimension as the common space (2048-dim in our case).
In terms of the semantics-guided contrastive learning subnet, we implement the MLP networks in RCCL, \emph{i.e.}, $g_s$ and $g_u$, as stacked linear layers with output size of [1024, 1]. Additionally, the MLP network in RRCL, \emph{i.e.}, $h_v$, is implemented as a single fully-connected layer with output size of 512. Except for the last layer of $g_u$ that uses Sigmoid activation, all the other linear layers are implemented with ReLU activation. We employ SGD optimizer with momentum of 0.9 and learning rate of 0.01 to optimize the proposed model. 

\noindent\textbf{Comparison Methods.}
To demonstrate the effectiveness of the proposed method, we compare it with both the baseline method and state-of-the-art approaches developed for the ZSD task. 
We provide a brief description of the comparison methods as follows.
\noindent \textbf{ConSE-ZSD} is the baseline method that adapts the standard Faster R-CNN model trained without any semantic information to the ZSD task by employing ConSE \cite{norouzi2013zero} at the testing stage.
\noindent \textbf{SAN} \cite{rahman2018zero} is the first deep network developed for the ZSD task that jointly models the interplay between visual and semantic domain information.
\noindent \textbf{HRE} \cite{demirel2018zero} is a YOLO \cite{redmon2017yolo9000} based end-to-end zero-shot detector that learns a direct mapping from region pixels to the space of class embeddings. 
\noindent \textbf{SB and DSES} are background-aware zero-shot detectors proposed in \cite{bansal2018zero} that differentiate background regions based on a large open vocabulary.
\noindent \textbf{TD} \cite{li2019zero} learns both visual-unit-level and word-level attention to tackle the ZSD task with textual descriptions instead of a single word.
\noindent \textbf{PL} \cite{rahman2018polarity} designs a novel polarity loss for RetinaNet based ZSD framework to better align visual and semantic concepts.
\noindent \textbf{BLC} \cite{zheng2020background} integrates Cascade Semantic R-CNN, semantic information flow and background learnable RPN into a unified ZSD framework.

\noindent\textbf{Evaluation Metrics.}
We adopt the widely-used evaluation protocols, \emph{i.e.}, Recall@100 and mAP, to evaluate the performance of our model, where a larger recall or mAP value indicates better performance \cite{bansal2018zero,rahman2018zero}. 
Specifically, Recall@100 is defined as the recall with only the top 100 detections from an image, while mAP indicates the mean average precision of the detection.
For mAP, we first calculate the per-class average precision (AP) for each individual class to study category-wise performance, then take the mean (mAP) as a measure of overall performance. 
More specifically, the widely adopted 11-point interpolation approach \cite{everingham2010pascal} is used to compute AP, which is defined as the average precision of eleven equally spaced recall levels [0, 0.1, 0.2, $\dots$, 1].
For ZSD, the testing phase only involves samples from unseen categories, and thus the performance is measured over the set of unseen classes $\mathcal{Y}^u$. While for GZSD, the samples from both seen categories $\mathcal{Y}^s$ and unseen categories $\mathcal{Y}^u$ are utilized to test the model performance.
The harmonic mean performance on seen and unseen classes is computed to reflect the overall performance for GZSD.

\subsection{Quantitative Results}
\subsubsection{PASCAL VOC}
 \begin{table}
 	\renewcommand{\arraystretch}{1.2}\setlength{\tabcolsep}{8pt}
	\centering
 	\begin{center}
 		\caption{ZSD and GZSD mAP(\%) at IoU threshold 0.5 on PASCAL VOC dataset, where ``S'' and ``U'' refer to the average performance on seen and unseen classes with ``HM'' denoting their harmonic mean.}
 		\label{voc_results1}
 		\begin{tabular}{lccccc}
 			\toprule
 			\multirow{2}*{Model}&\multirow{2}*{Seen}&\multirow{2}*{ZSD}&\multicolumn{3}{c}{GZSD}\\ \cmidrule{4-6}
 			~&&&S&U&HM\\
 			\midrule
 			ConSE-ZSD&\textbf{77.0}&52.1&59.3&22.3&32.4\\
 			SAN& 69.6&59.1&48.0&37.0&41.8\\
 			HRE &65.6&54.2&62.4&25.5&36.2\\
 			PL&63.5&62.1&-&-&-\\
 			BLC&75.1&55.2&58.2&22.9&32.9\\
 			\textbf{ContrastZSD}&76.7&\textbf{65.7}&\textbf{63.2}&\textbf{46.5}&\textbf{53.6}\\
 			\bottomrule
 		\end{tabular}	
 	\end{center}
 \end{table} 
\noindent\textbf{ZSD and GZSD Performance.}
 \begin{table*}[t]
 	\renewcommand{\arraystretch}{1.2}
 	\tabcolsep=3.5 pt	
 	\begin{center}
 		\caption{Class-wise AP and mAP (\%) on the PASCAL VOC dataset at IoU threshold 0.5, where mAP$_s$ and mAP$_u$ refer to the mAP values with respect to seen and unseen classes respectively.}\label{voc_results2}
 		\begin{tabular}{l|ccccccccccccccccc|ccccc}
 			\toprule
 			Methods& \rotatebox{90}{\emph{aeroplane}} & \rotatebox{90}{\emph{bicycle}} & \rotatebox{90}{\emph{bird}} & \rotatebox{90}{\emph{boat}} & \rotatebox{90}{\emph{bottle}}&\rotatebox{90}{\emph{bus}}&\rotatebox{90}{\emph{cat}}&\rotatebox{90}{\emph{chair}}&\rotatebox{90}{\emph{cow}}&\rotatebox{90}{\emph{d. table}} &\rotatebox{90}{\emph{horse}}&\rotatebox{90}{\emph{motorbike}}&\rotatebox{90}{\emph{person}} &\rotatebox{90}{\emph{p. plant}}&\rotatebox{90}{\emph{sheep}}&\rotatebox{90}{\emph{tvmonitor}}&\rotatebox{90}{mAP$_s$}&\rotatebox{90}{\emph{car}} &\rotatebox{90}{\emph{dog}}&\rotatebox{90}{\emph{sofa}}&\rotatebox{90}{\emph{train}} &\rotatebox{90}{mAP$_u$}\\
 			\midrule
 			~&\multicolumn{17}{c|}{Seen Classes}&\multicolumn{5}{c}{Unseen Classes}\\
 			\midrule
 			ConSE-ZSD&\textbf{82.2}&\textbf{85.8}&83.2&66.7&70.0&77.5&87.4&\textbf{60.1}&\textbf{80.0}&69.4&84.5&\textbf{85.0}&84.6&\textbf{56.6}&\textbf{81.4}&78.1& \textbf{77.0}&49.0&75.0&53.0&31.3&52.1\\
 			SAN &71.4&78.5&74.9&61.4&48.2&76.0&\textbf{89.1}&51.1&78.4&61.6&84.2&76.8&76.9&42.5&71.0&71.7& 69.6&56.2&85.3&62.6&26.4&57.6\\
 			HRE &70.0&73.0&76.0&54.0&42.0&\textbf{86.0}&64.0&40.0&54.0&\textbf{75.0}&80.0&80.0&75.0&34.0&69.0&\textbf{79.0}& 65.6&55.0&82.0&55.0&26.0&54.2\\
 			PL &74.4 &71.2 &67.0 &50.1 &50.8 &67.6 &84.7 &44.8 &68.6 &39.6& 74.9 &76.0 &79.5 &39.6 &61.6 &66.1& 63.5&63.7 &\textbf{87.2} &53.2 &44.1& 62.1\\
 			BLC &78.5 &83.2 &77.6 &\textbf{67.7} &70.1 &75.6 &87.4 &55.9 &77.5 &71.2& \textbf{85.2} &82.8 &77.6 &56.1 &77.1 &78.5& 75.1&43.7 &86.0 &60.8 &30.1& 55.2\\
 			\textbf{ContrastZSD}&81.9& 85.6& \textbf{85.0}&66.6& \textbf{70.8}& 77.0& 88.9&58.4& 79.5&66.8& 84.7& 82.2& \textbf{84.9}& 55.4& 81.1& 78.4&76.7& \textbf{65.5}& 86.4& \textbf{63.1}& \textbf{47.9}& \textbf{65.7}\\
 			\bottomrule
 		\end{tabular}	
 	\end{center}
 \end{table*}
We present the mAP performance in Table \ref{voc_results1} to compare different methods over the PASCAL VOC dataset. Based on the settings in \cite{demirel2018zero}, the performance of each method is reported in three different testing configurations, \emph{i.e.}, ``Seen'', ``ZSD'' and ``GZSD'', where ``Seen'' refers to the conventional object detection task used to detect objects from $\mathcal{Y}^s$. 
We can observe from Table \ref{voc_results1} that our method outperforms all the comparison methods under the ``ZSD'' setting, increasing the mAP from 62.1\% achieved by the second-best method PL to 65.7\%, which indicates the good transferable ability of the proposed ContrastZSD model to unseen classes.
In addition to the state-of-the-art ZSD performance, it's interesting to see that our model also performs well on the general ``Seen'' object detection task, even comparable to the ConSE-ZSD baseline that only focuses on training an excellent Faster R-CNN model over seen classes.
We can attribute this good performance to the region-region contrastive learning subnet of our model that optimizes the visual structure for better discriminating different seen classes. Despite the effectiveness of ConSE-ZSD on seen object detection, it achieves the worst performance on ZSD task due to the lack of semantic information in the training phase. 

In contrast to the ``Seen'' and ``ZSD'' task, ``GZSD'' is a more challenging  and realistic task where both seen and unseen classes are present at inference.
As depicted in Table \ref{voc_results1}, for each comparison method, the unseen object detection performance of GZSD drops significantly compared with that of the corresponding ZSD results. {One possible reason for this performance degradation is that those comparison methods can easily bias towards the seen classes, such that most unseen objects are recognized as seen classes during testing.}
Compared with those methods, our model shows more promising results on unseen object detection for GZSD, \emph{i.e.}, 46.5\% \emph{vs} 37.0\%, while not disturbing the seen object detection performance, \emph{i.e.}, 63.2\% \emph{vs} 62.4\%. Thus, our method can obtain a more balanced performance on the seen and unseen classes for GZSD.

\noindent\textbf{Class-wise Performance.}
To study the per-category results, we present the class-wise mAP performance on PASCAL VOC in Table \ref{voc_results2}. The results on seen and unseen classes are evaluated independently in ``Seen'' and ``ZSD'' setting for fair comparison with other methods. Not surprisingly, the baseline method ConSE-ZSD shows more promising results on seen classes than other methods. As shown in Table \ref{voc_results2}, ConSE-ZSD achieves the best performance on 7 out of 16 seen categories, \emph{e.g.}, ``aeroplane'', ``chair'' and ``cow''. As for the class-wise ZSD results, our method outperforms all the competitors on three of the four unseen classes by a large margin, which further verifies the superiority of our model for the ZSD task. Compared with other methods, the performance gain is more pronounced for ``car'' and ''train'' classes. We think this is because the car and train objects are visually similar, making the system hard to distinguish. Benefiting from the region-region contrastive learning strategy, our model can learn more discriminative visual features to better distinguish objects belonging to the two categories.

\subsubsection{MS COCO}
  \begin{table}
 	\renewcommand{\arraystretch}{1.2}
 	\tabcolsep=5.5 pt	
 	\begin{center}
 		\caption{ZSD performance in terms of Recall@100(\%) and mAP(\%) with different IoU thresholds on MS COCO dataset.}\label{coco zsd}
 		\begin{tabular}{lccccc}
 			\toprule	
 			\multirow{2}*{Model}&\multirow{2}*{Split}&\multicolumn{3}{c}{Recall@100}&mAP\\ \cmidrule{3-6}
 			~&&IoU=0.4&IoU=0.5&IoU=0.6&IoU=0.5\\
 			\midrule	
 			ConSE-ZSD&48/17&28.0&19.6&8.7&3.2\\
 			SB&48/17&34.5 &22.1 &11.3 &0.3\\
 			DSES&48/17&40.2& 27.2& 13.6 &0.5\\
 			TD&48/17&45.5 &34.3 &18.1&-\\
 			PL&48/17&-&43.5&-&10.1\\
 			BLC&48/17&51.3 &48.8 &45.0 &10.6\\
 			\textbf{ContrastZSD}&48/17&\textbf{56.1}&\textbf{52.4}&\textbf{47.2}&\textbf{12.5}\\
 			\midrule
 			ConSE-ZSD&65/15&30.4&23.5&10.1&3.9\\
 			PL&65/15&-&37.7&-&12.4\\
 			BLC&65/15&57.2 &54.7 &51.2 &14.7\\
 			\textbf{ContrastZSD}&65/15&\textbf{62.3}&\textbf{59.5}&\textbf{55.1}&\textbf{18.6}\\
 			\bottomrule
 		\end{tabular}	
 	\end{center}
 \end{table} 
\begin{table}
	\renewcommand{\arraystretch}{1.2}
	\tabcolsep= 6pt	
	\begin{center}
		\caption{GZSD performance in terms of Recall@100 (\%) and mAP (\%) achieved with IoU=0.5 over each seen/unseen split of MS COCO.}\label{coco gzsd}
		\begin{tabular}{lccccccc}
			\toprule
			\multirow{2}*{Method}&\multirow{2}*{Split}&\multicolumn{3}{c}{Recall@100}&\multicolumn{3}{c}{mAP}\\
            \cmidrule{3-8}
			&&S&U&HM&S&U&HM\\
			\midrule
			ConSE-ZSD& 48/17&43.8&12.3&19.2&37.2&1.2&2.3\\
			PL& 48/17&38.2&26.3&31.2&35.9&4.1&7.4\\
			BLC& 48/17&57.6&46.4&51.4&42.1&4.5&8.2\\
			\textbf{ContrastZSD}& 48/17&\textbf{65.7}&\textbf{52.4}&\textbf{58.3}&\textbf{45.1}&\textbf{6.3}&\textbf{11.1}\\
			\midrule
			ConSE-ZSD& 65/15 &41.0&15.6&22.6&35.8&3.5&6.4\\
			PL& 65/15&36.4&37.2&36.8&34.1&12.4&18.2\\
			BLC& 65/15&56.4&51.7&53.9&36.0&13.1&19.2\\
			\textbf{ContrastZSD}& 65/15 &\textbf{62.9}&\textbf{58.6}&\textbf{60.7}&\textbf{40.2}&\textbf{16.5}&\textbf{23.4}\\
			\bottomrule
		\end{tabular}
	\label{one}	
	\end{center}
\end{table}
\begin{table*}
\caption{Class-wise Recall@100 for the 48/17 and 65/15 split of MS-COCO with the IoU threshold being 0.5.}\label{coco classwise}
    \begin{minipage}{1\linewidth}
    \renewcommand{\arraystretch}{1.2}
      \centering
 		\begin{tabular}{lcccccccccccccccccc}
 			\toprule
 			48/17 split& \rotatebox{90}{\emph{bus}} & \rotatebox{90}{\emph{dog}} & \rotatebox{90}{\emph{cow}} & \rotatebox{90}{\emph{elephant}} & \rotatebox{90}{\emph{umbrella}}&\rotatebox{90}{\emph{tie}}&\rotatebox{90}{\emph{skateboard}}&\rotatebox{90}{\emph{cup}}&\rotatebox{90}{\emph{knife}}&\rotatebox{90}{\emph{cake}} &\rotatebox{90}{\emph{couch}}&\rotatebox{90}{\emph{keyboard}}&\rotatebox{90}{\emph{sink}} &\rotatebox{90}{\emph{scissors}}&\rotatebox{90}{\emph{airplane}}&\rotatebox{90}{\emph{cat}}&\rotatebox{90}{\emph{snowboard}} &\rotatebox{90}{mean(\%)}\\
 			\midrule
 			BLC-Base& 72.9& \textbf{94.6}& 67.3& 68.1& 0.0& 0.0& 19.9& 24.0& 12.4& 24.0& 63.7& 11.6& 9.2& 8.3& 48.3& 70.7& 63.4&38.7\\
 			BLC&77.4& 88.4 &71.9 &77.2 &0.0& 0.0 &41.7 &38.0 &\textbf{45.6} &34.3 &65.2 &23.8 &14.1 &20.8 &48.3 &79.9 &61.8& 46.4\\
 			\textbf{ContrastZSD}&\textbf{82.8}&92.1& \textbf{76.9}& \textbf{82.0}& \textbf{2.3}& \textbf{1.1}& \textbf{45.0}& \textbf{51.7}& 41.7&\textbf{44.2}& \textbf{74.2}& \textbf{33.7}& \textbf{21.0}& \textbf{32.3}& \textbf{55.6}&\textbf{83.8}& \textbf{69.5}& \textbf{52.4}\\
 			\bottomrule
 		\end{tabular}
    \end{minipage}
    \begin{minipage}{1\linewidth}
    \renewcommand{\arraystretch}{1.2}
 	\tabcolsep=7 pt	
      \centering
 		\begin{tabular}{lcccccccccccccccc}
 		\multicolumn{17}{c}{}\\
 			\toprule
 			65/15 split& \rotatebox{90}{\emph{airplane}} & \rotatebox{90}{\emph{train}} & \rotatebox{90}{\emph{p. meter}} & \rotatebox{90}{\emph{cat}} & \rotatebox{90}{\emph{bear}}&\rotatebox{90}{\emph{suitcase}}&\rotatebox{90}{\emph{frisbee}}&\rotatebox{90}{\emph{snowboard}}&\rotatebox{90}{\emph{fork}}&\rotatebox{90}{\emph{sandwich}} &\rotatebox{90}{\emph{h. dog}}&\rotatebox{90}{\emph{toilet}}&\rotatebox{90}{\emph{mouse}} &\rotatebox{90}{\emph{toaster}}&\rotatebox{90}{\emph{h. drier}}&\rotatebox{90}{mean(\%)}\\
 			\midrule
 			 BLC-Base&53.9& 70.6& 5.9& 90.2& 85.1& 40.7& 25.9& 59.9& 33.7& 76.9& 64.4& 33.2& 3.3& \textbf{64.1} &1.4&47.3\\
 			BLC&58.7& 72.0& 10.2&96.1& 91.6 & 46.9& 44.1& 65.4& 37.9 &82.5& \textbf{73.6}& 43.8& 7.9& 35.9& 2.7& 51.3\\
 			\textbf{ContrastZSD}&\textbf{67.7}& \textbf{77.5}& \textbf{17.3}& \textbf{97.4}& \textbf{94.6}& \textbf{56.6}& \textbf{57.2}& \textbf{72.0}& \textbf{43.7}&\textbf{85.0}& \textbf{73.6}& \textbf{67.7}& \textbf{17.6}&47.4& \textbf{4.1}& \textbf{58.6}\\
 			\bottomrule
 		\end{tabular}	
    \end{minipage}
  \end{table*}

\noindent\textbf{ZSD Performance.}
For the MS COCO dataset, we follow the experimental settings in \cite{bansal2018zero} and \cite{li2019zero} to evaluate the ZSD performance with different IoU thresholds, \emph{i.e.}, 0.4, 0.5 and 0.6.
The experimental results in terms of both Recall@100 and mAP are presented in Table \ref{coco zsd}. For the 48/17 split, we compare our model with ConSE-ZSD, SB, DSES, TD, PL and BLC. From the ZSD results in Table \ref{coco zsd}, we can observe that our method achieves a significant gain on both metrics (mAP and Recall@100). Compared with the second-best method BLC, the proposed model gains an absolute improvement of 1.9\% in mAP and 3.6\% in Recall@100 at IoU threshold 0.5.
On the 65/15 split, we compare our model only with ConSE-ZSD, PL and BLC, since other methods didn't report their results on this split.
As shown in Table \ref{coco zsd}, the proposed model outperforms all the comparison methods by a large margin, which improves the mAP and Recall@100 achieved by the second-best method BLC from 14.7\% and 54.7\% to 18.6\% and 59.5\% at IoU threshold 0.5. {These improvements demonstrates the effectiveness and significance of the proposed contrastive model on detecting unseen objects.}

\noindent\textbf{GZSD Performance.}
In Table \ref{coco gzsd}, we further present the GZSD results achieved by ConSE-ZSD, PL, BLC and the proposed ContrastZSD. The results demonstrate that our model exceeds the three comparison methods in terms of both mAP and Recall@100. As shown in Table \ref{coco gzsd}, the proposed ContrastZSD outperforms the second-best method BLC by a large margin, where the absolute HM performance gain is 6.9\% Recall@100 and 2.9\% mAP for the 48/17 split and 6.8\% Recall@100 and 4.2\% mAP for the 65/15 split. Moreover, it is worth noting that the performance gain of our model is more remarkable on GZSD than ZSD, as shown in Table \ref{coco zsd} and \ref{coco gzsd}. {This phenomenon reflects the significance of the proposed explicit knowledge transfer on the GZSD task.}
Due to the lack of semantic information during model training, the performance of ConSE-ZSD is far worse than the other methods on both of mAP and Recall@100 metrics. {In contrast, PL yeilds notable improvement over ConSE-ZSD by adopting the direct mapping-transfer strategy. However, the recall and mAP on seen classes are much higher than on unseen classes, leading to a low harmonic mean (HM) performance.} This is because the mapping-transfer strategy is prone to over-fitting the seen classes, such that very little knowledge is learned for unseen classes.

\noindent\textbf{Class-wise Performance.}
The class-wise performance on unseen classes of the two splits is reported in Table \ref{coco classwise} under the GZSD setting. Compared with the state-of-the-art BLC method, our model achieves higher Recall@100 on 15 out of 17 unseen classes on 48/17 split and 14 out of 15 unseen classes on 65/15 split. This phenomenon suggests that our model can improve the GZSD performance evenly, instead of only focusing on certain categories. 
We have also noted that BLC fails to detect any objects from the ``umbrella'' and ``tie'' class, resulting in a recall rate of 0.
One possible reason is that those classes have fewer semantically similar concepts in the seen category set, which greatly increases the difficulty of knowledge transfer. {Benefiting from the region-category contrastive learning over unseen classes, the training proposals from a group of related seen classes are explicitly leveraged to train the unseen classes.}
As a result, our model successfully surpasses BLC over those unseen classes without close counterparts in the seen class set, \emph{e.g.}, the ``umbrella'', ``tie'' and ``hair drier'' class.

\subsection{Ablation Studies}
\subsubsection{Ablation for Contrastive Learning Subnets}
We conduct extensive quantitative analysis for the key components, \emph{i.e.}, RRCL and RCCL, in the proposed model by leaving one component out of our framework at a time. In table \ref{ablation}, we present the ZSD and GZSD performance in terms of mAP on the PASCAL VOC dataset to compare the effects of different contrastive learning subnets. The results of ``ContrastZSD'' are obtained by simultaneously considering all the components, leading to the best ZSD and GZSD performance. 
 \begin{table}
 	\renewcommand{\arraystretch}{1.2}
 	\tabcolsep=5 pt	
 	\begin{center}
 		\caption{The effect of each key component for ZSD and GZSD performance in terms of mAP at IoU threshold 0.5 over PASCAL VOC dataset.}\label{ablation}
 		\begin{tabular}{lcccccc}
 			\toprule
 			\multirow{2}*{Variants}&\multirow{2}*{RCCL$_u$}&\multirow{2}*{RRCL}&\multirow{2}*{ZSD}&\multicolumn{3}{c}{GZSD}\\
 			\cmidrule{5-7}
 			&&&&S&U&HM\\
 			\midrule	
 			w/o. RRCL&$\surd$&&61.5&59.3&44.2&50.6\\
 			w/o. RCCL$_u$&&$\surd$&61.2&61.0&30.6&40.8\\
 			\textbf{ContrastZSD}&$\surd$&$\surd$&\textbf{65.7}&\textbf{63.2}&\textbf{46.5}&\textbf{53.6}\\
 			\bottomrule
 		\end{tabular}	
 	\end{center}
 \end{table} 
  \begin{figure}
	\centering 	
	\subfigure[Varying $\lambda$]{\label{4a}	
		\includegraphics[width=0.48\linewidth]{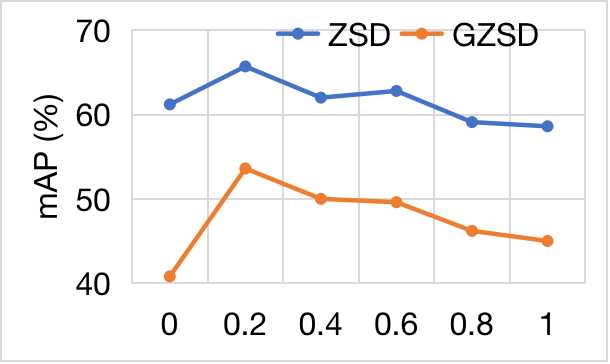}}
	\subfigure[Varying $\beta$]{\label{4b}		
		\includegraphics[width=0.48\linewidth]{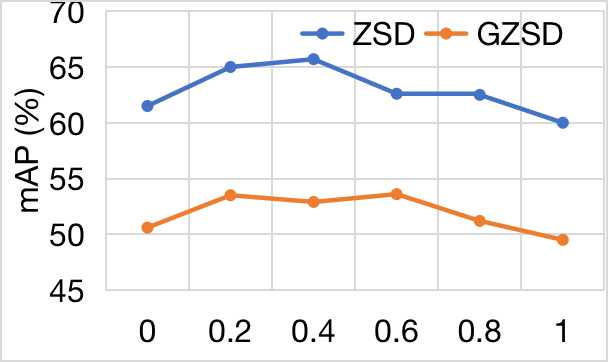}}
	\caption{Hyper-parameter sensitivity of the proposed ContrastZSD model on the PASCAL VOC dataset. 
	}	
	\label{sensitivity}	
\end{figure}
\begin{figure*}[t]
	\centering 	
	\subfigure[Seen visual features produced by ConSE-ZSD]{\label{5a}		
		\includegraphics[width=0.48\linewidth]{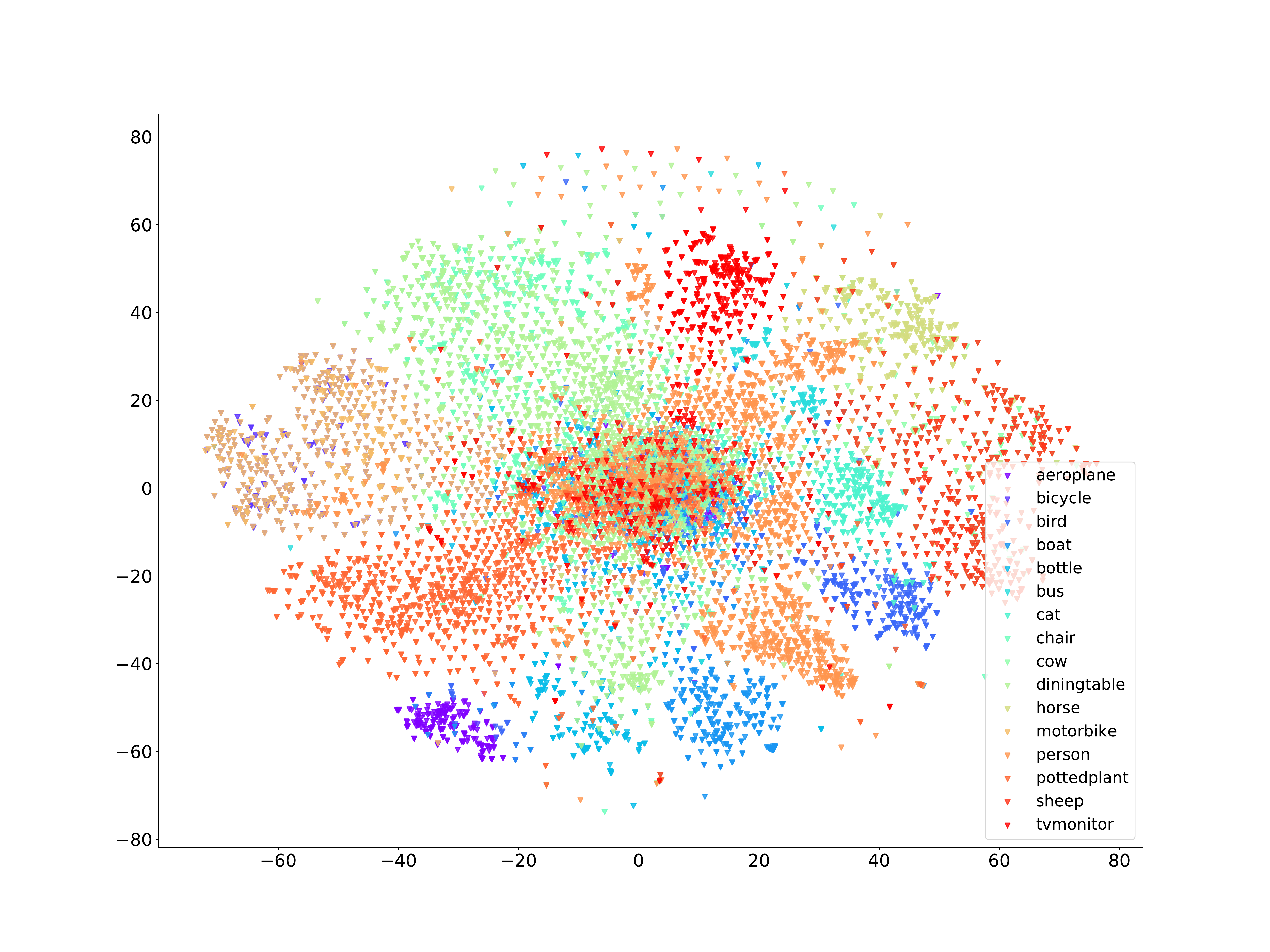}}\hspace{10pt}
	\subfigure[Seen visual features produced by ContrastZSD]{\label{5b}			
		\includegraphics[width=0.48\linewidth]{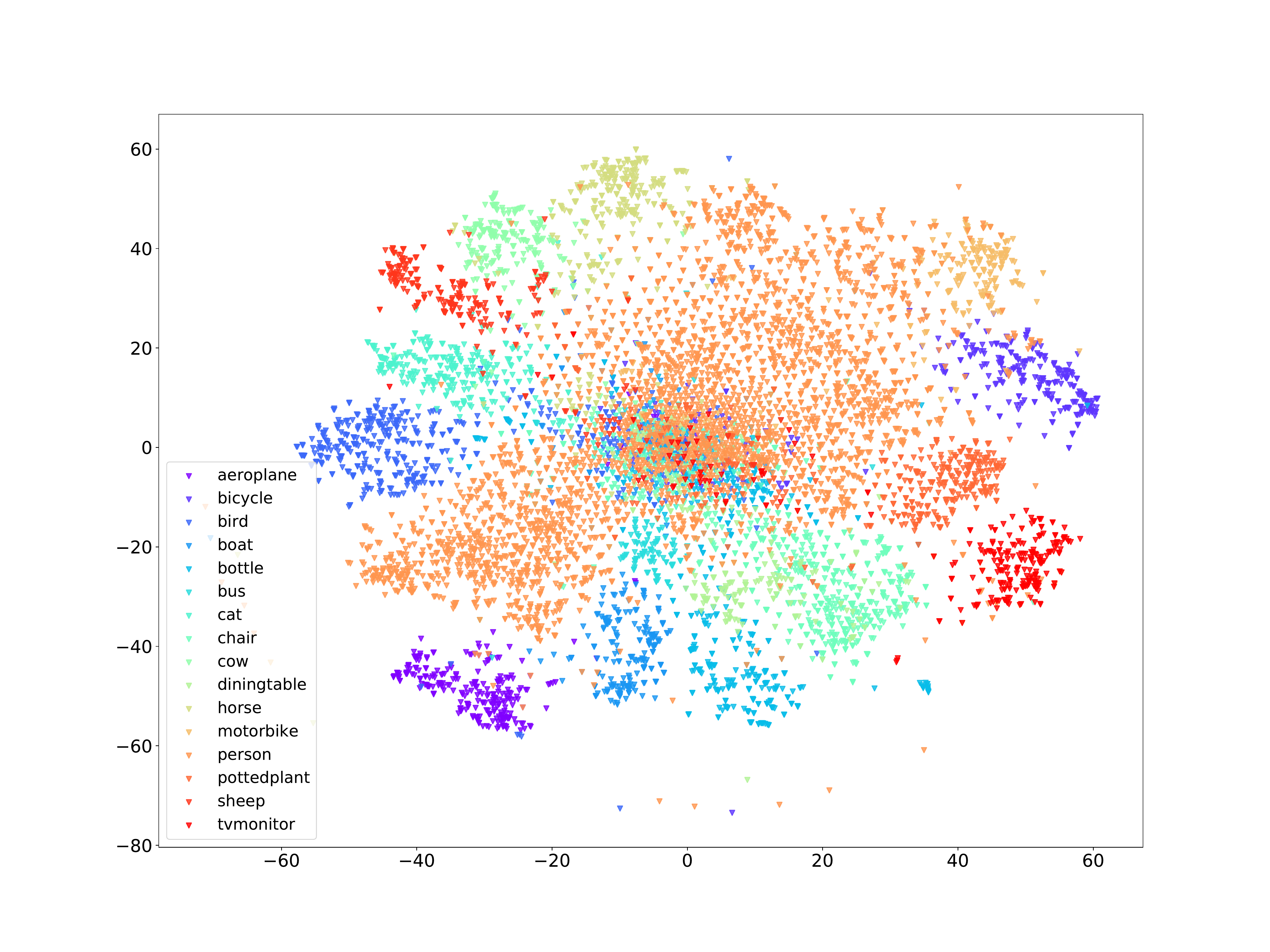}}
	\vfill
	\subfigure[Uneen visual features produced by ConSE-ZSD]{\label{5c}	
		\includegraphics[width=0.48\linewidth]{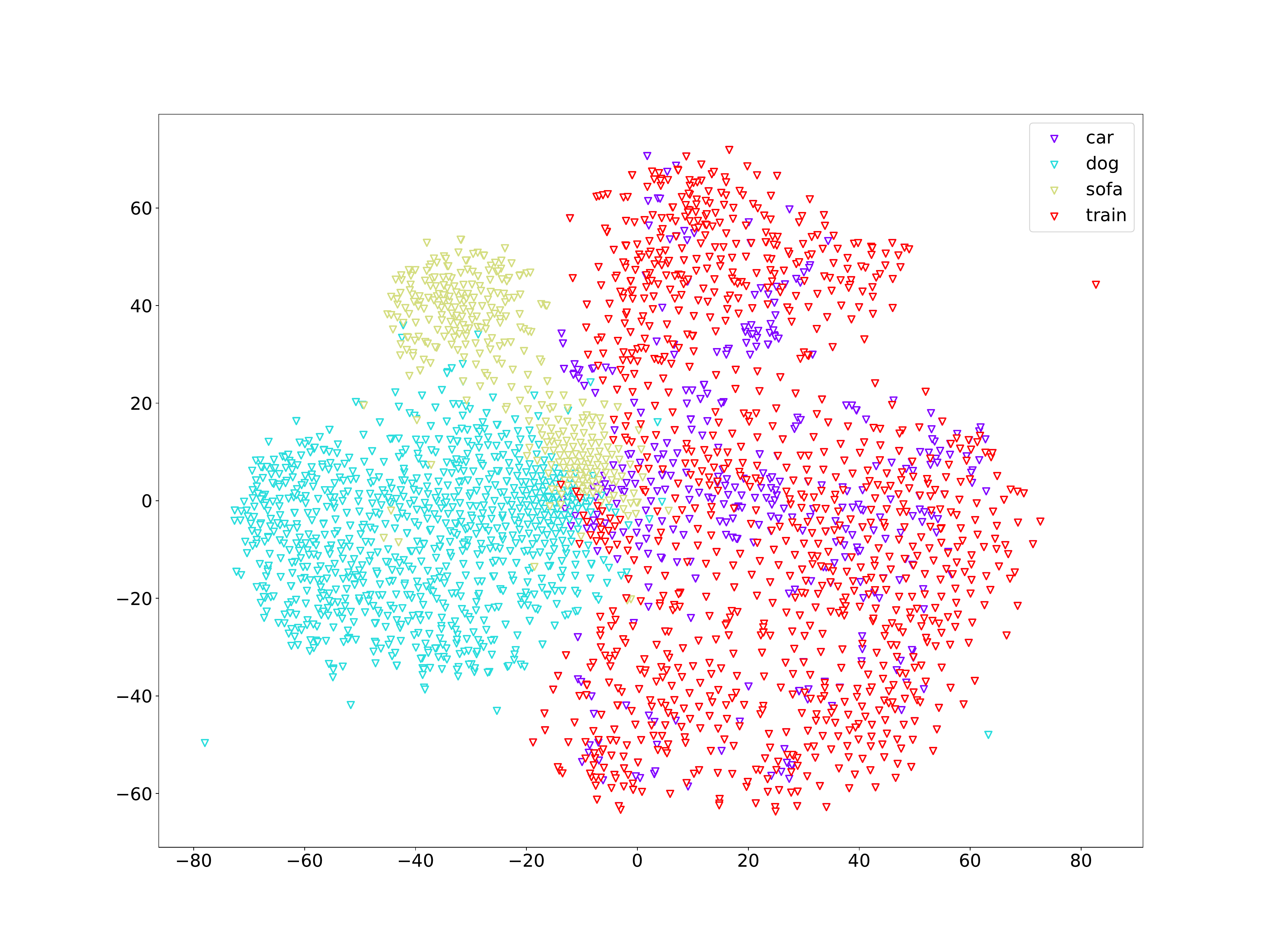}}\hspace{10pt}
	\subfigure[Unseen visual features produced by ContrastZSD]{\label{5d}		
		\includegraphics[width=0.48\linewidth]{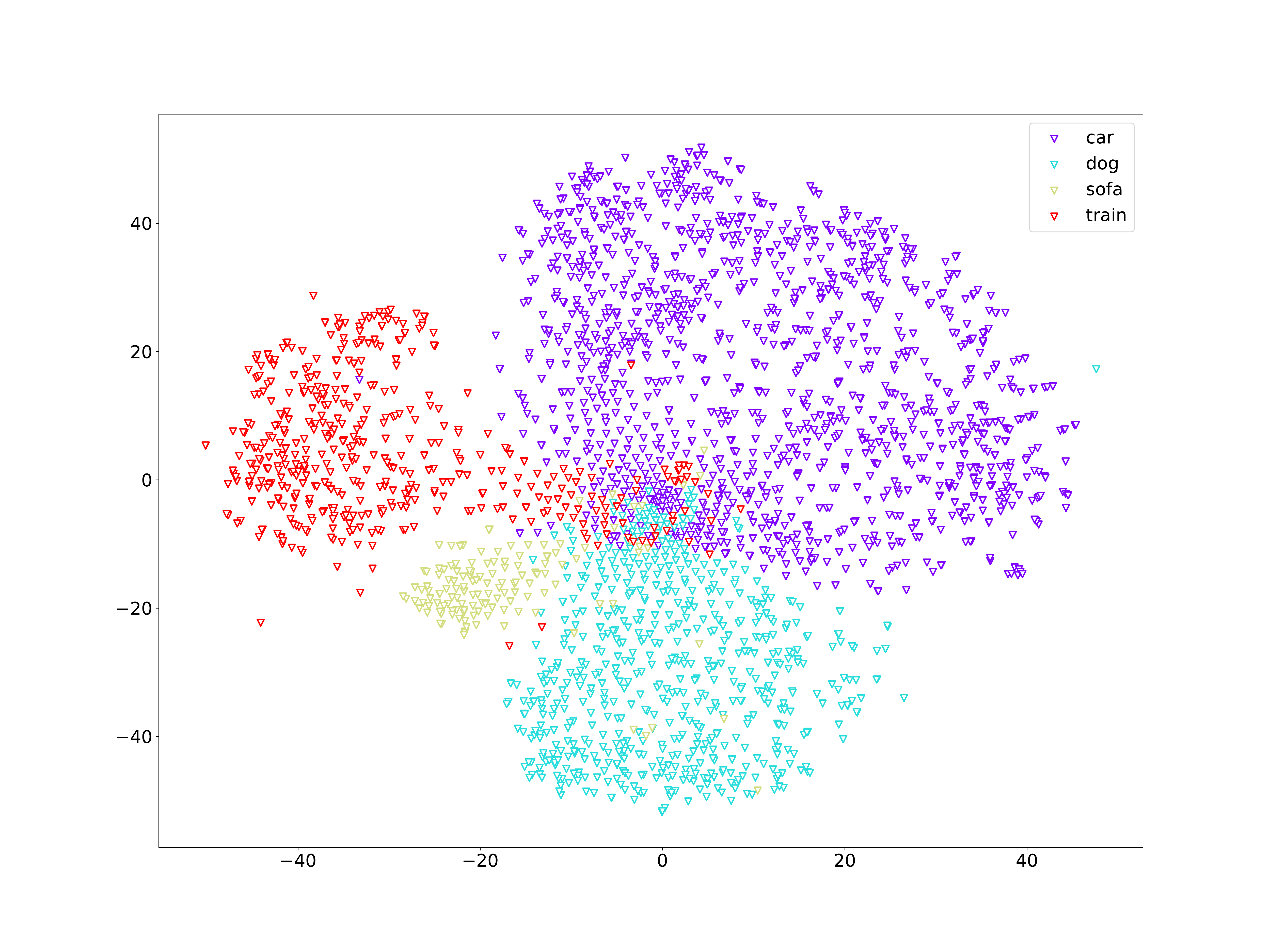}}
	\caption{Visualization of the visual feature distribution on PASCAL VOC using t-SNE, where the points from different categories are marked in different colors. The visual features of seen and unseen classes are shown in (a) (b) and (c) (d) respectively. }	
	\label{t-sne-pascal}	
\end{figure*}
\begin{figure*}[t] 	
	\centering 		
	\includegraphics[width=1\linewidth]{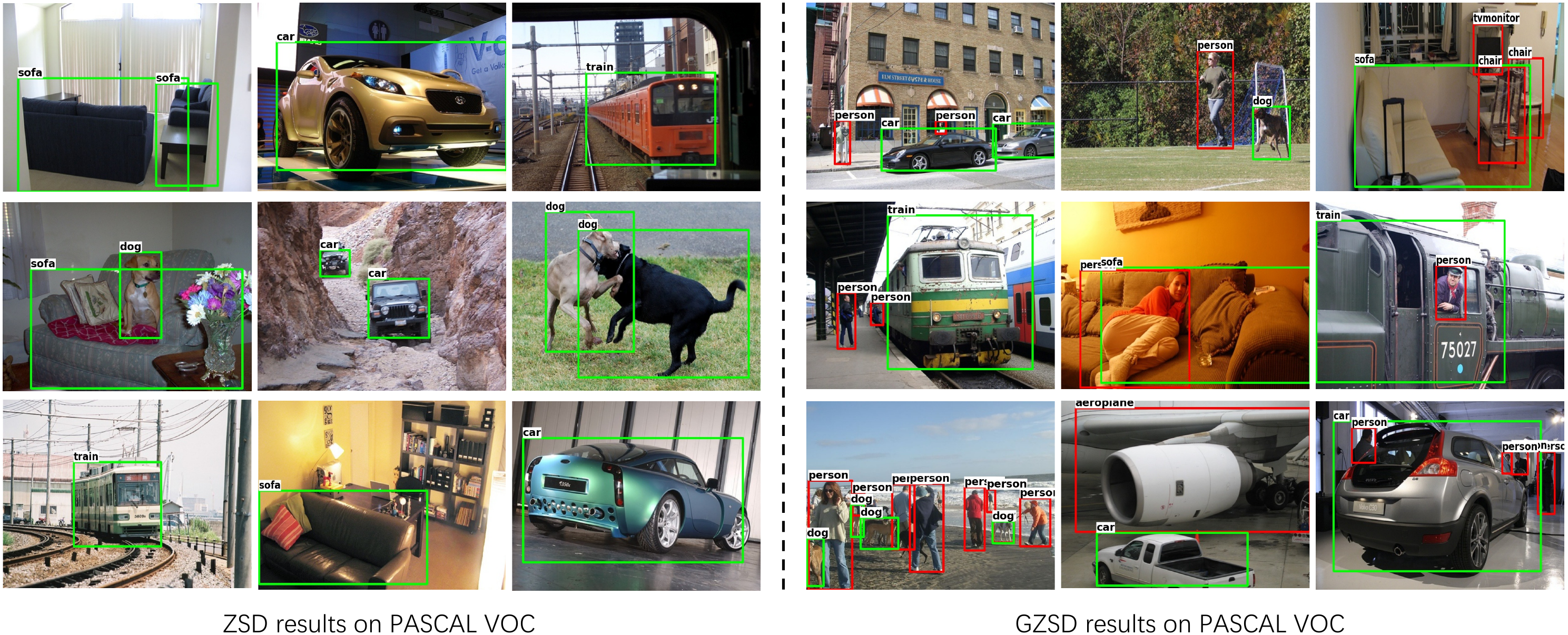}	
	\caption{Some ZSD and GZSD detection results on the PASCAL VOC dataset. The region proposals of seen and unseen categories are marked as red and green boxes respectively.}
	\label{pascal_visualize}	
\end{figure*}
\begin{figure*}[t] 	
	\centering 		
	\includegraphics[width=0.9\linewidth]{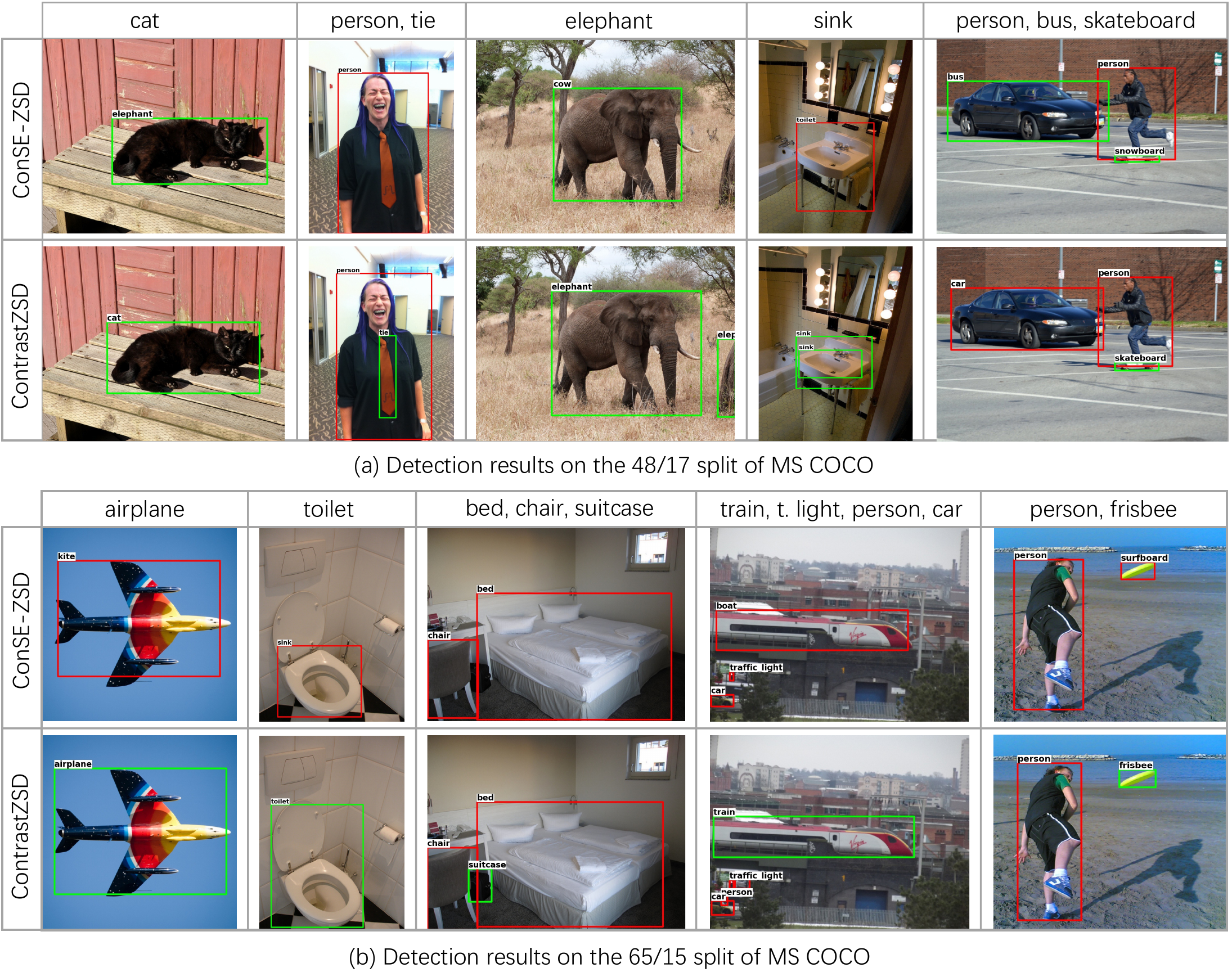}
	\caption{Some ZSD and GZSD detection results on two splits of the MS COCO dataset. For each split, the detection results in the first and second row are produced by ConSE-ZSD and ContrastZSD respectively.}
	\label{coco_visualize}	
\end{figure*}

\noindent\textbf{Effectiveness of RCCL$_u$.}
The method ``w/o. RCCL$_u$'' removes the unseen class contrastive learning process in the RCCL subnet to contrast seen objects merely with seen categories, thus the explicit knowledge transfer from seen to unseen classes cannot be conducted. As a result, both the ZSD and GZSD mAP suffer from a degradation compared with ContrastZSD. Notably, while there is only a small decrease on the ZSD performance, the mAP on unseen classes of GZSD drops significantly from 48.3\% to 30.6\%, leading to a low harmonic mean performance. This phenomenon indicates the explicit knowledge transfer plays a more important role in GZSD than ZSD, since it can prevent the model from biasing towards the seen classes.

\noindent\textbf{Effectiveness of RRCL.}
``w/o. RRCL'' denotes the variant method that removes the RRCL subnet, such that the original visual space cannot be optimized based on the class label information. Compare with ContrastZSD, the mAP performance over both ZSD and GZSD experiences a decline, \emph{i.e.}, 61.5\% \emph{vs} 65.7\% on ZSD and 50.6\% \emph{vs} 53.6\% on GZSD. This is because the original visual space is lack of discriminative ability and thus is suboptimal for ZSD/GZSD.
With the help of the RRCL subnet, our model can optimize the visual data structure, including both the seen and unseen distribution, to be more distinguishable.
\subsubsection{Sensitivity Analysis}
In order to investigate the importance of each key component, we further analyze the effect of hyper-parameters to our model by varying $\lambda$ and $\beta$ in the range of $\{$0, 0.2, 0.4, 0.6, 0.8, 1$\}$. The ZSD and GZSD performance in terms of mAP achieved with varying parameters on PASCAL VOC are demonstrated in Fig. \ref{sensitivity}.

\noindent\textbf{Sensitivity Analysis for $\lambda$.} We first discuss the impact of parameter $\lambda$ on the performance of the proposed ContrastZSD. As shown in Fig. \ref{4a}, when the value of $\lambda$ increases from 0, the performance of our model gains a notable improvement. This indicates that the explicit knowledge transfer in RCCL can indeed enable the model to learn more knowledge about the unseen domain. 
Notably, choosing $\lambda$ around 0.2 tends to yield the best ZSD and GZSD performance. If we keep increases the value of $\lambda$, both of the ZSD and GZSD performance begin to decrease. Thus, we conjecture that $\lambda$ should be set small in order to achieve good performance. 

\noindent\textbf{Sensitivity Analysis for $\beta$.} Then we discuss the impact of the parameter $\beta$ on our model that controls the contribution of the RRCL subnet. 
As shown in Fig. \ref{4b}, the best choice of $\beta$ is 0.4 for ZSD and 0.6 for GZSD respectively over the PASCAL VOC dataset. Larger or smaller values of parameter $\beta$ tend to degrade the detection performance.
It proves that the visual structure constraint in RRCL subnet can effectively optimize the visual data distribution to be more distinguishable with proper $\beta$, allowing for better visual-semantic alignment. Taking both ZSD and GZSD into consideration, we set $\beta$ to 0.5 in our experiments.

\subsection{Qualitative Analysis}
\noindent\textbf{Visual Structure Optimization.}
To further demonstrate the effectiveness of our model on visual structure optimization, we utilize t-SNE \cite{maaten2008visualizing} to visualize the visual features of detected region proposals on the PASCAL VOC dataset. 
The visual features produced by the baseline method ConSE-ZSD are illustrated in Fig. \ref{5a} and \ref{5c}, where most clusters of the categories fail to have a clear frontier.
For example, the intra-class distance of the ``horse'' and ``sheep'' objects in Fig. \ref{5a} is sometimes even larger than their inter-class distance, while the ``car'' and ``train'' class objects in Fig. \ref{5c} suffer from an extremely large overlap.
In such scenarios, the objects from different classes are extremely hard to be distinguished, thereby significantly inhibiting the learning of embedding functions.
By contrast, it can be clearly observed from Fig. \ref{5b} and \ref{5d} that the visual features learned by our model demonstrate higher intra-class compactness, as well as a much larger inter-class margin on both the seen and unseen categories of PASCAL VOC, exhibiting more obvious clustering patterns.
This phenomenon verifies that our model is able to produce more discriminative visual features for better visual-semantic alignment, which further substantiates the above-mentioned quantitative improvements.

\noindent\textbf{Detection Results.}
For qualitative analysis of the detection performance, we present some ZSD and GZSD results on PASCAL VOC and MS COCO dataset in Fig. \ref{pascal_visualize}
 and Fig. \ref{coco_visualize} respectively. From the ZSD results on PASCAL VOC shown in Fig. \ref{pascal_visualize}(a), we can figure out that our model is capable of detecting unseen objects under different scenarios: (a) a single object in an image, \emph{e.g.}, ``car'', ``train'' and ``sofa''; (b) multiple objects from the same category, \emph{e.g.}, ``car'' and ``dog''; (c) multiple objects from different categories, \emph{e.g.}, ``sofa'' and ``dog'.
 Besides, we have also noted that our model is capable of detecting objects from both seen and unseen classes in the same image, as depicted in Fig. \ref{pascal_visualize}(b).
 For example, $\{$``car'', ``person''$\}$, $\{$``sofa'', ``chair'', ``tvmonitor''$\}$ and $\{$``dog'', ``pottedplant'', ``tvmonitor''$\}$ are detected in the same image respectively, where ``car'', ``sofa'' and ``dog'' are unseen objects.
 These examples confirm that the proposed model can be applied successfully to both the ZSD and GZSD tasks. 
 For MS COCO, we show qualitative comparison between our model and the baseline method ConSE-ZSD, both of which are based on the Faster R-CNN framework. From Fig. \ref{coco_visualize}, it's interesting to see that ConSE-ZSD can localize the bounding box for most of the objects from either seen or unseen classes, although it did not use any semantic information during training. We can attribute this to the good generalization ability of the region proposal network in Faster R-CNN that generates objects in an objectness manner. However, ConSE-ZSD fails to predict the true class label for most of the unseen objects. For example, ConSE-ZSD recognizes the ``elephant'' object as ``cow'' in Fig. \ref{coco_visualize}(a), and ``airplane'' object as ``kite'' in Fig. \ref{coco_visualize}(b), \emph{etc}. By contrast, our method provides more accurate detection results for either seen or unseen objects in the selected images. Moreover, our model also successfully detects the objects that have been missed by ConSE-ZSD, like the ``tie'' object in Fig. \ref{coco_visualize}(a) and ``suitcase'' object in Fig. \ref{coco_visualize}(b).

\section{Conclusion}
In this paper, we have made the first attempt to facilitate the zero-shot object detection task with contrastive learning mechanism, and developed a novel ContrastZSD framework for ZSD.
{The proposed ContrastZSD incorporates two contrastive learning subnets guided by semantics information, \emph{i.e.}, RCCL and RRCL, in order to guarantee both the discriminative and transferable property.}
Specifically, the RRCL subnet optimizes the visual data distribution in the joint embedding space to be more distinguishable based on class label information.
The RCCL subnet enables explicit knowledge transfer from seen classes to unseen classes, {thereby alleviating the model's bias problem towards seen classes.}
The quantitative and qualitative experimental results confirm that the proposed framework improves the performance of both the ZSD and GZSD task.


%



\section*{Acknowledgments}

This study was supported by the National Key Research and Development Program of China (No. 2020AAA0108800), National Nature Science Foundation of China (No. 61872287, No. 62050194, No. 61922064, No. 61922064 and U2033210), Innovative Research Group of the National Natural Science Foundation of China (No. 61721002),  Innovation Research Team of Ministry of Education (IRT\_17R86), Project of China Knowledge Center for Engineering Science and Technology, and Project of Chinese academy of engineering ``The Online and Offline Mixed Educational Service System for `The Belt and Road' Training in MOOC China'', and the Zhejiang Provincial Natural Science Foundation (No. LR17F030001). Prof Chang was partially supported by Australian Research Council (ARC) Discovery Early Career Researcher Award (DECRA) under DE190100626.

\ifCLASSOPTIONcaptionsoff
  \newpage
\fi



%



\bibliographystyle{IEEEtran}
\bibliography{./ref}

%
\begin{IEEEbiography}[{\includegraphics[width=1in,height=1.25in,clip,keepaspectratio]{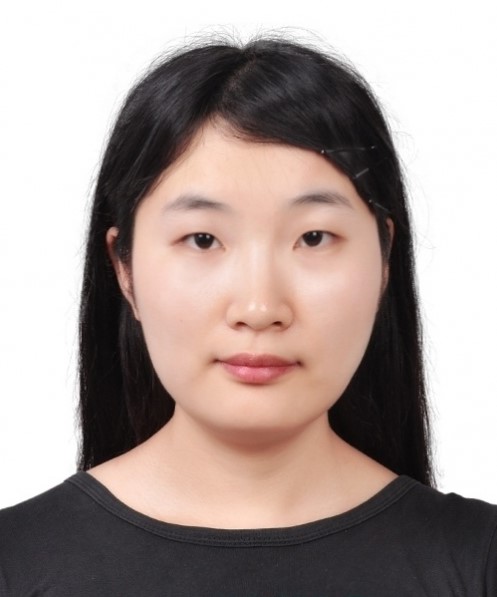}}]{Caixia Yan} received the BS degree in Computer Science and technology from Xi'an Jiaotong University in 2015. She is currently working toward the Ph.D. degree in computer science and technology at Xi'an Jiaotong University. She was also a visiting scholar in the School of Computer Science at Carnegie Mellon University. Her research interests include machine learning and optimization, multiple feature learning and image processing.
\end{IEEEbiography}

\begin{IEEEbiography}[{\includegraphics[width=1in,height=1.85in,clip,keepaspectratio]{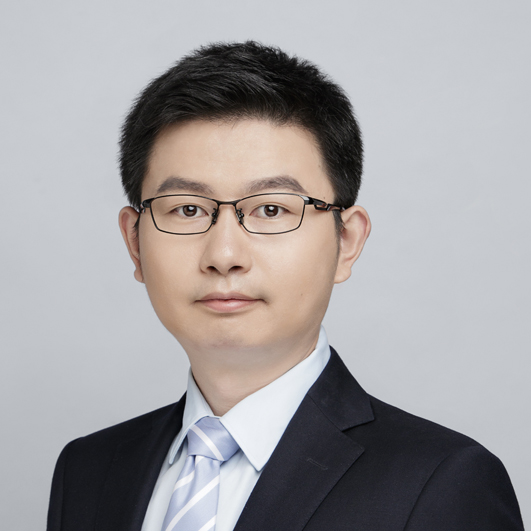}}] {Xiaojun Chang} is a Professor at Faculty of Engineering and Information Technology, University of Technology Sydney.
He was an ARC Discovery Early Career Researcher Award (DECRA) Fellow between 2019-2021. After graduation, he was worked as a Postdoc Research Associate in School of Computer Science, Carnegie Mellon University, a Senior Lecturer in Faculty of Information Technology, Monash University, and an Associate Professor in School of Computing Technologies, RMIT University. He mainly worked on exploring multiple signals for automatic content analysis in unconstrained or surveillance videos and has achieved top performance in various international competitions. He received his Ph.D. degree from University of Technology Sydney. His research focus in this period was mainly on developing machine learning algorithms and applying them to multimedia analysis and computer vision.
\end{IEEEbiography}


\begin{IEEEbiography}[{\includegraphics[width=1in,height=1.25in,clip,keepaspectratio]{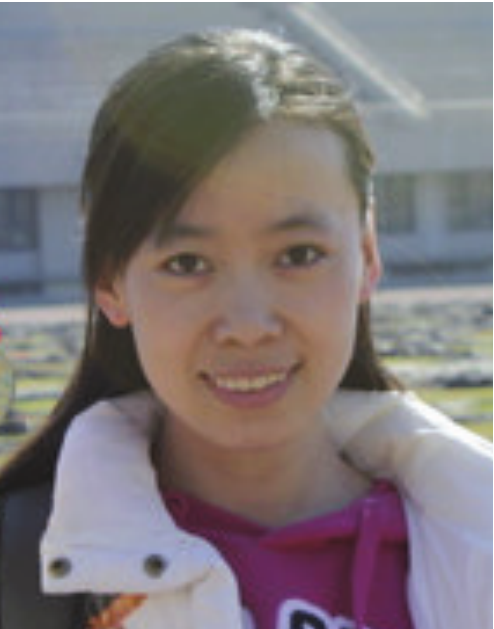}}]{Minnan Luo}
	received the Ph. D. degree from the Department of Computer Science and Technology, Tsinghua University, China, in 2014. Currently, she is an Assistant Professor in the School of Electronic and Information Engineering at Xi'an Jiaotong University. She was a Post-Doctoral Research with the School of Computer Science, Carnegie Mellon University, Pittsburgh, PA, USA.
	Her research interests include machine learning and optimization, cross-media retrieval and fuzzy system.
\end{IEEEbiography}

\begin{IEEEbiography}[{\includegraphics[width=1in,height=1.25in,clip,keepaspectratio]{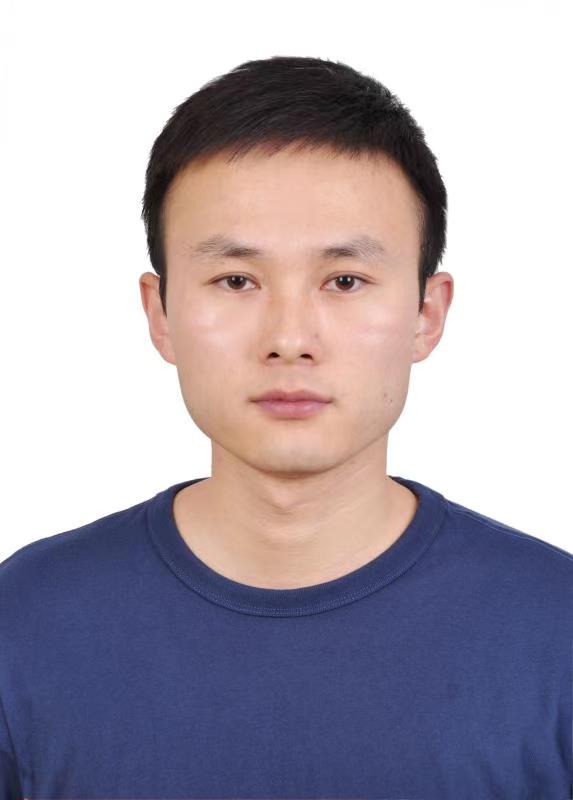}}]{Huan Liu} received the B.S. and Ph.D. degrees in computer science from Xi'an Jiaotong University, China, in 2013 and 2020, respectively. He is currently an assistant professor at Xi'an Jiaotong University. His research areas include deep learning, machine learning, and their application in computer vision and EEG-based affective computing.
\end{IEEEbiography}

\begin{IEEEbiography}[{\includegraphics[width=1in,height=1.25in,clip,keepaspectratio]{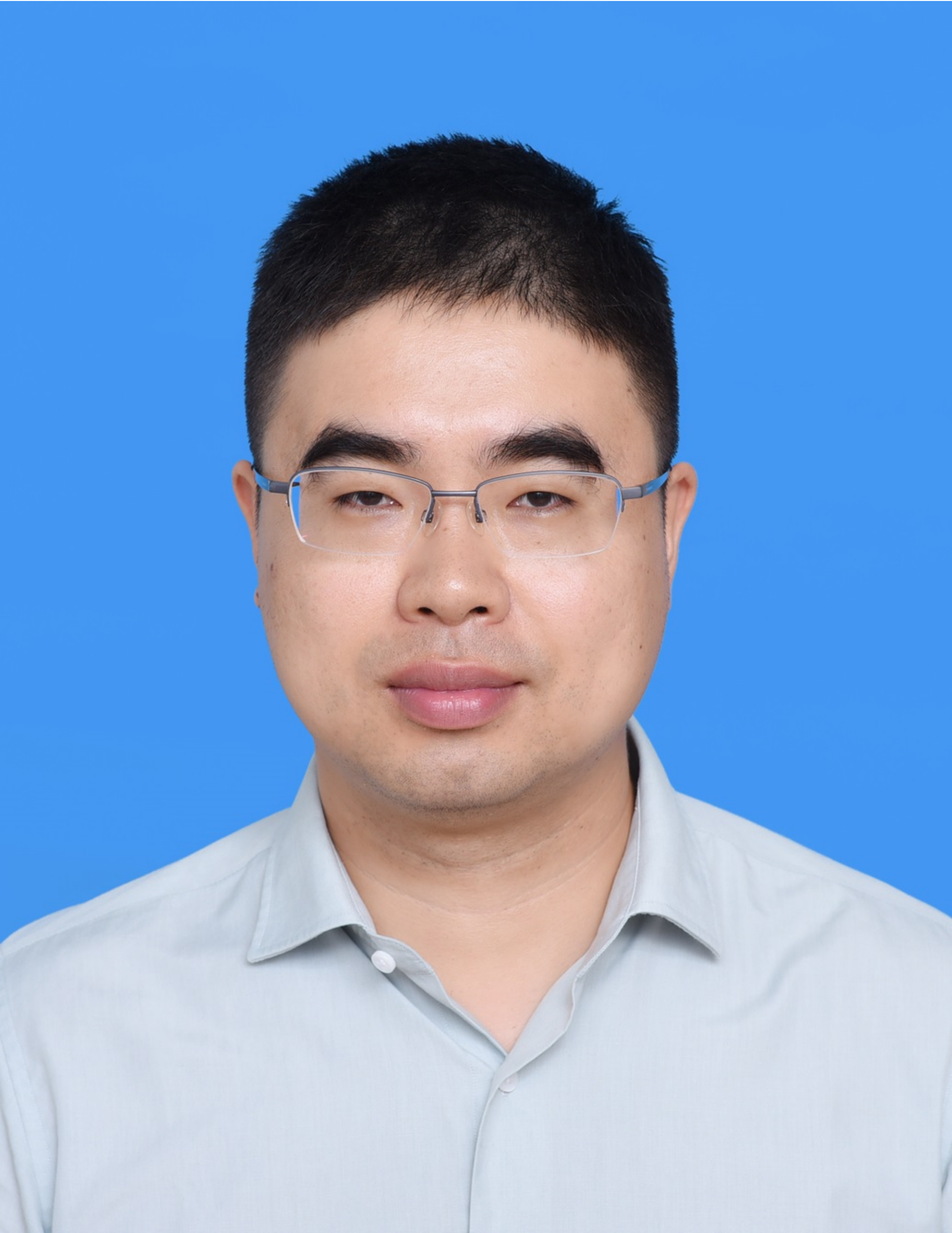}}]{Xiaoqin Zhang}
 received the B.Sc. degree in electronic information science and technology from Central South University, China, in 2005, and the Ph.D. degree in pattern recognition and intelligent system from the National Laboratory of Pattern Recognition, Institute of Automation, Chinese Academy of Sciences, China, in 2010. \\ He is currently a professor with Wenzhou University, China. He has authored or co-authored over 100 papers in international and national journals and international conferences. His research interests are in pattern recognition, computer vision, and machine learning.
\end{IEEEbiography}

\begin{IEEEbiography}[{\includegraphics[width=1in,height=1.25in,clip,keepaspectratio]{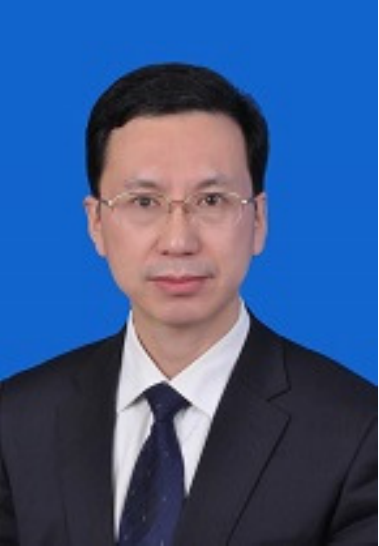}}]{Qinghua Zheng}
	received the B.S. degree in computer software in 1990, the M.S. degree in computer organization and architecture in 1993, and the Ph.D. degree in system engineering in 1997 from Xi'an Jiaotong University, China. He was a postdoctoral researcher at Harvard University in 2002. He is currently a professor in Xi'an Jiaotong University. His research areas include computer network security, intelligent E-learning theory and algorithm, multimedia e-learning, and trustworthy software.
\end{IEEEbiography}




\end{document}